\documentclass{article} %
\usepackage{iclr2027_conference,times}

\usepackage{algorithm}
\usepackage{algpseudocode}
\usepackage{multirow}
\usepackage{graphicx}
\usepackage{subcaption}

\usepackage[utf8]{inputenc} %
\usepackage[T1]{fontenc}    %

\usepackage{booktabs}       %
\usepackage{amsfonts}       %
\usepackage{nicefrac}       %
\usepackage{microtype}      %

\usepackage{amsmath,amsfonts,bm}

\def\eqref#1{equation~\ref{#1}}

\def\1{\bm{1}}

\DeclareMathAlphabet{\mathsfit}{\encodingdefault}{\sfdefault}{m}{sl}
\SetMathAlphabet{\mathsfit}{bold}{\encodingdefault}{\sfdefault}{bx}{n}

\usepackage{hyperref}
\usepackage{url}
\usepackage{xcolor}

\renewcommand{\eqref}[1]{(\ref{#1})}

\title{Learning While Searching: Inverse Design through Online Refinement and Offline Reuse}

\author{Mohanad Obeed \& Ming Jian \\
Huawei Technologies Canada \\
Ottawa, ON, Canada}

\iclrfinalcopy
\begin{document}

\maketitle
\lhead{Preprint}

\begin{abstract}
Inverse design seeks physical parameters that produce a specified response; it is
hard when the mapping is nonlinear or non-unique and each simulation is expensive.
We introduce ORO (Online Refinement, Offline Reuse), which learns while it searches by
coupling a forward surrogate with a target-conditioned inverse design network. Online,
the inverse network proposes a design for the target; the simulated response corrects
the surrogate, and gradients through it move the inverse network toward the target,
with neither simulator derivatives nor an acquisition function. Offline, every
accumulated design--response pair retrains the surrogate at no extra simulation cost,
and the inverse network is retrained through it; a decreasing fraction of calls explore
from the best design found so far. With every call counted and all methods searching
the same bounds, ORO attains the lowest median error at $2{,}800$ calls per target
against a genetic algorithm, CMA-ES, Bayesian optimization, TuRBO and AutoTandemML.
On an ill-conditioned Rosenbrock-valley benchmark its median error of
$1.3\times10^{-6}$ is $10$ times below CMA-ES, the strongest baseline there ($25/30$
paired wins), $129$ times below TuRBO and at least $1{,}900$ times below the other
baselines, each on all $30$ targets. On a resonance benchmark it is $45$ times below
TuRBO, the strongest baseline there ($26/30$), and on a pole--zero benchmark it is level
with TuRBO ($16/30$) and beats all other methods; TuRBO leads at small budgets when each
target starts from scratch. Offline
retraining is what makes the surrogate's gradients useful, reducing the true error on
$92\%$ and $79\%$ of steps against $47\%$ without it, and training the inverse network
offline on the requested target rather than on achieved responses helps on the
resonance benchmark. Designing $30$ targets in sequence from a task-agnostic warm-up,
ORO keeps the lowest median error on all three benchmarks, $282$ times below TuRBO on
Rosenbrock ($30/30$) and $20$ times below on resonance ($28/30$), level with it on the
pole--zero benchmark, and on Rosenbrock later targets reach lower error at every
budget. On a band-pass filter designed with a
commercial full-wave solver, ORO reaches a compliant design in about $1{,}200$
simulations, against $16{,}300$ for a surrogate-based genetic approach.
\end{abstract}

\section{Introduction}

Engineering design often begins with a desired behavior: an optical spectrum, a
radio-frequency S-parameter response, a mechanical deformation, or a fluid-flow
characteristic. Given a geometry, material configuration, and operating
conditions, established solvers can predict the resulting response. This forward
problem is well defined, even when its numerical solution is expensive. Inverse
design reverses the task: it seeks physical parameters that produce a specified
response, a formulation shared across electromagnetics, photonics, mechanics,
and fluid dynamics \citep{Lalau-Keraly:13, ICLR2024_7f8b8bc8}.

Three difficulties make this substantially harder than forward prediction.
The inverse relation is often non-unique: different configurations may produce
indistinguishable responses, so a network trained to reproduce individual design
labels faces conflicting supervision even though any of several designs would
satisfy the specification. Response-based evaluation is therefore more
appropriate than requiring recovery of a particular parameter vector
\citep{liu2018training}. Even a uniquely solvable problem can be nonlinear,
nonconvex, and poorly conditioned, with narrow curved valleys and sensitivities
differing by orders of magnitude. Finally, physical and manufacturing
constraints restrict feasible designs, and a requested response may not be
exactly attainable; the practical objective is an admissible design with
sufficiently small response error.

These challenges compound when the forward model is an expensive black box. A
high-fidelity simulation may take minutes or hours, limiting both optimization
and training-data generation, and broad sampling wastes much of that budget on
responses far from the specification. Learning introduces a further difficulty:
designs proposed during optimization can leave the region covered by the
training data, where a surrogate may predict an attractive response for a design
that performs poorly under the true simulator---a well-known failure mode in
model-based optimization \citep{pmlr-v139-trabucco21a}. The central problem is
thus to coordinate candidate selection and model learning under a limited
evaluation budget.

Classical optimization repeatedly generates and evaluates candidates. Genetic
algorithms search complex design spaces without simulator gradients and have long been
applied to photonic and electromagnetic design \citep{10.1117/12.545902}, but they
reduce each simulated response curve to a single fitness value, discarding most of the
acquired information, and their evaluations do not transfer to the next specification.
Adjoint optimization computes many sensitivities from few field solves, but requires
sensitivity information unavailable in the black-box setting considered here.

Surrogate-assisted methods instead learn an approximation of the forward process
and optimize through it, as in the neural-adjoint method
\citep{NEURIPS2020_007ff380}. A fixed surrogate, however, inherits the coverage of
its training set, which must be generated before the specification is known:
accuracy is distributed uniformly while the search concentrates in one region.
Adaptive sampling addresses the choice of evaluations explicitly. Bayesian
optimization maintains a probabilistic model and selects queries through an
acquisition rule balancing predicted performance and uncertainty
\citep{NIPS2012_05311655, eriksson2019scalable}, and AutoTandemML combines
predictive-uncertainty sampling with tandem networks \citep{GRBCIC2025113828}.
Related ideas appear in experimental design for inverse problems, where PIED
optimizes measurement designs through differentiable physics-informed networks
\citep{hemachandra2025pied}, and in generative inverse models such as conditional
invertible networks, which represent the full distribution of designs consistent
with a response \citep{ardizzone2019analyzing,ardizzone2019guided}; both assume
either a known governing equation or a fixed training set rather than a black-box
simulator queried during the search.
These methods reuse observations effectively, but select new evaluations through
a separate criterion that must itself be maximized over the design space at
every round, and that criterion is target-agnostic: it seeks where the forward
model is uncertain, not where the requested design lies. Physics-driven
generative optimization couples generation with learning more directly
\citep{jiang2019global}, but obtains its training gradients from adjoint
simulations.

\paragraph{Proposed approach.} We propose \emph{Online Refinement, Offline
Reuse} (ORO), a framework that learns while it searches. It maintains a
differentiable forward surrogate $S_\phi$, an inverse design network $D_\theta$,
and an accumulating dataset of simulator observations. During online refinement,
the inverse network is fed the target performance and generates
$x_t = D_\theta(y^{\star})$; the simulator returns $y_t = F(x_t)$. This single
observation first updates the surrogate to better predict the proposed design's
response. With the surrogate then held fixed, the inverse network is updated by
differentiating $\ell(S_\phi(D_\theta(y^{\star})), y^{\star})$. The simulator
supplies ground-truth responses; the surrogate supplies the approximate
gradients that move the next proposal.

This arrangement has a consequence that distinguishes ORO from the
adaptive-sampling methods above: \emph{no acquisition function exists}.
The next query is the forward pass of the target-conditioned network, which moves
only because $\theta$ moved under the target-error gradient: there is no utility to
design, no exploration--exploitation weighting to tune and no inner maximization, and,
unlike \citet{jiang2019global}, only response evaluations are required. Because this
refinement is local, ORO adds a simple scheduled exploration---random single-parameter
moves from the best design found so far---which is a mutation operator rather than an
acquisition criterion (Section~\ref{sec:explore}).

Periodic offline phases then reuse every accumulated design--response pair: the
surrogate is retrained on the whole dataset, and the inverse network is retrained
through it, at no additional simulator cost. What the inverse network should be
trained on offline is a design choice we study explicitly. Each achieved response
$y_i$ is a solved problem whose design is known; the requested target $y^{\star}$
is the problem actually being solved. We find that the
requested target performs better on one benchmark and equally on the other, and
that the benefit of the offline phase comes mainly from retraining the surrogate,
which is what makes its gradients point downhill on the true error
(Section~\ref{sec:mechanism}). Each ingredient of ORO has a precedent: training an
inverse model through a learned forward model \citep{jordan1992forward}, the neural
adjoint \citep{NEURIPS2020_007ff380}, tandem networks with active learning
\citep{GRBCIC2025113828}, hindsight relabelling \citep{NIPS2017_453fadbd} and goal
babbling \citep{rolf2010goal}. Goal babbling is the closest: it also runs a
goal-conditioned inverse model toward requested goals and learns from the outcomes it
reaches, but fits the inverse model directly to those outcomes, weighting samples to
resolve redundancy, in robot kinematics where trials are cheap. ORO instead learns
through a differentiable forward surrogate, which supplies a gradient toward a target
not yet reached, under a fixed budget of expensive simulations. Our contribution is the
integration, in which one target-conditioned network is both the design model and the
query rule, together with an analysis of which part produces the gains.

In summary, this paper makes the following contributions:
\begin{itemize}
\item \textbf{A framework in which acquisition is not a separate component.} A
target-conditioned inverse network, trained through an adaptively corrected forward
surrogate, serves as both the design model and the query rule, with no acquisition
criterion and no simulator derivatives; a fixed exploration schedule complements the
local refinement.

\item \textbf{Controlled evaluation against strong baselines.} With every call
counted, identical bounds and reported wall-clock costs, ORO attains the lowest median
error at the full budget against GA, CMA-ES, BO, TuRBO and AutoTandemML. On the
ill-conditioned Rosenbrock benchmark it is $10$ times below CMA-ES and $129$ times below
TuRBO, and better than TuRBO, GA, BO and AutoTandemML on all $30$ targets and than
CMA-ES on $25$; it is $45$ times below TuRBO on resonance and level with TuRBO on the
pole--zero benchmark.

\item \textbf{An analysis of what drives the method.} Gradient-alignment diagnostics,
component ablations, a comparison of offline training objectives and a comparison with
direct design optimization through the same surrogate show that offline retraining is
what makes the surrogate's gradients useful.

\item \textbf{Benchmarks and application.} A benchmark suite that separates
non-uniqueness from ill-conditioning; a sequential study in which ORO carries its data
across $30$ targets; and a band-pass filter designed with a commercial full-wave solver in about
$1{,}200$ simulations.
\end{itemize}

\section{Method}
\label{sec:method}

\subsection{Problem setup}
\label{sec:setup}

Let $F:\mathcal{X}\rightarrow\mathcal{Y}$ denote the simulator, mapping an
admissible design $x\in\mathcal{X}\subset\mathbb{R}^{d}$, where $d$ is the number
of design parameters, to a response $y=F(x)\in\mathbb{R}^{m}$ sampled at $m$
frequency (or spatial) points. Given a target response $y^{\star}$, the inverse
design objective is
\begin{equation}
x^{\star}\in\arg\min_{x\in\mathcal{X}}\ \ell\!\left(F(x),y^{\star}\right),
\qquad \text{where} \ 
\ell(a,b)=\tfrac{1}{m}\|a-b\|_{2}^{2}.
\label{eq:problem}
\end{equation}
We assume the weakest practical form of simulator access: $F$ is a deterministic
black box returning only $y=F(x)$. No derivatives, adjoint sensitivities, or
solver internals are available, and $F$ dominates cost, so methods are compared
primarily by the number of calls to $F$; Section~\ref{sec:budget_matched} also reports
wall-clock time.
Two structural properties of \eqref{eq:problem} shape the method. The solution is
generally non-unique: distinct designs may satisfy $F(x_{1})=F(x_{2})$, so
supervision in parameter space is ill-posed and all losses below are defined in
response space. And $y^{\star}$ need not be exactly attainable, so the objective
is an acceptable design with sufficiently small response error rather than an
exact inverse.

\subsection{Networks}
\label{sec:networks}

ORO maintains two networks. The \textbf{forward surrogate}
$S_{\phi}:\mathbb{R}^{d}\rightarrow\mathbb{R}^{m}$ approximates $F$; being
differentiable, it supplies the gradients that $F$ cannot. The \textbf{inverse
design network} $D_{\theta}:\mathbb{R}^{m}\rightarrow\mathcal{X}$ maps a desired
response to design parameters. Because it is conditioned on the response, a
design is obtained in a single forward pass and no search is performed at
inference. Admissibility is enforced in the output layer rather than by
constrained optimization: each output passes through a sigmoid scaled to its
interval of the search box $\mathcal{X}$, so every proposal satisfies
$D_{\theta}(\cdot)\in\mathcal{X}$ by construction, and $\mathcal{X}$ is the same box
searched by every baseline (Appendix~\ref{app:arch}). Both networks are multilayer
perceptrons; architectures are given in Appendix~\ref{app:arch}, and the
framework is agnostic to these choices.

Both phases below follow the same update pattern: the surrogate is fitted first,
then frozen while the inverse network is fitted through it. Gradients pass
through $S_{\phi}$ but never modify it during an inverse-network update.

\begin{figure}[t]
\centering
\begin{subfigure}[b]{0.33\textwidth}
\includegraphics[width=\linewidth]{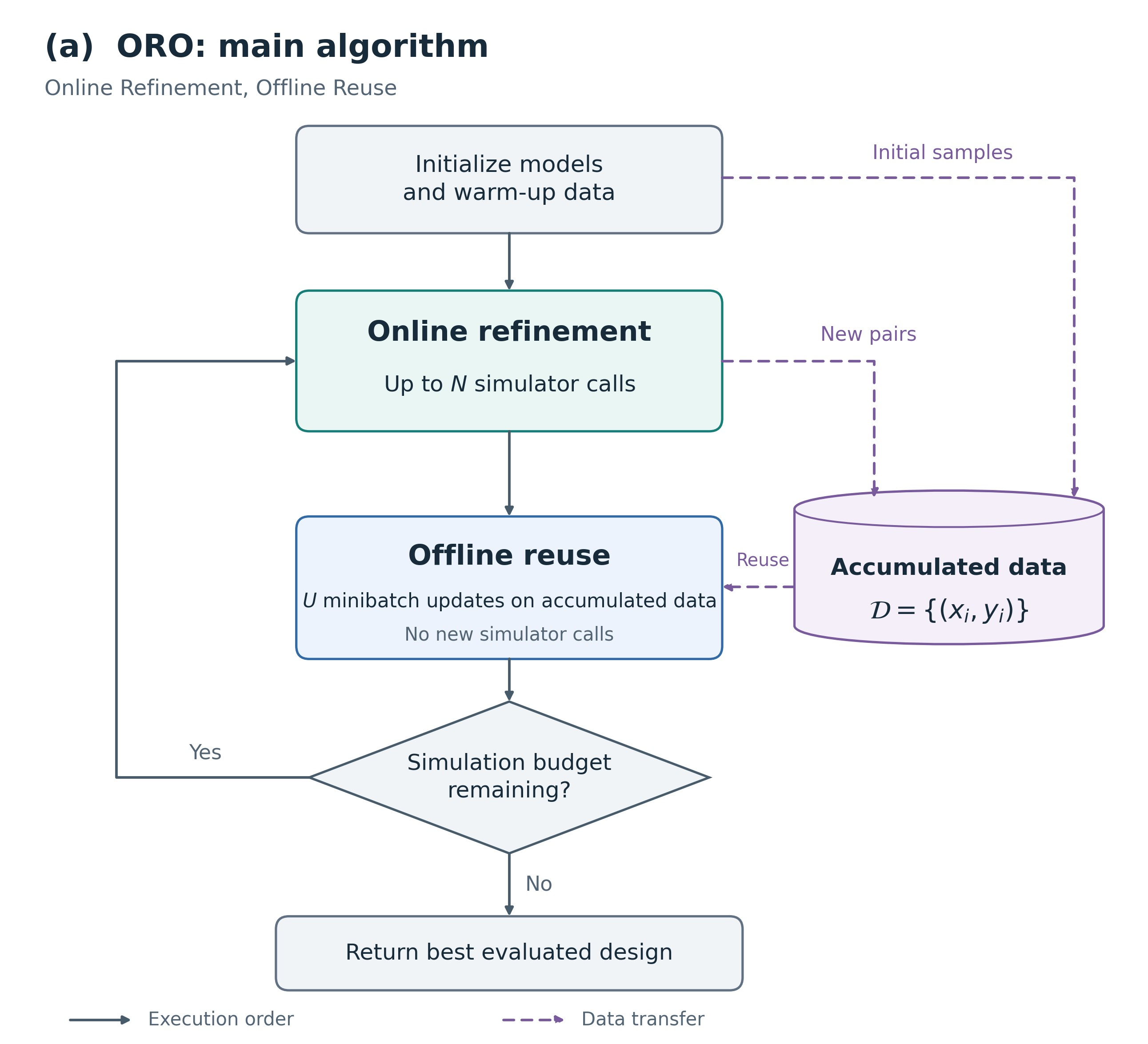}
  \caption{Main Algorithm}
  \label{fig:flow_main}
  
\end{subfigure}
\hfill
\begin{subfigure}[b]{0.32\textwidth}
\includegraphics[width=\linewidth]{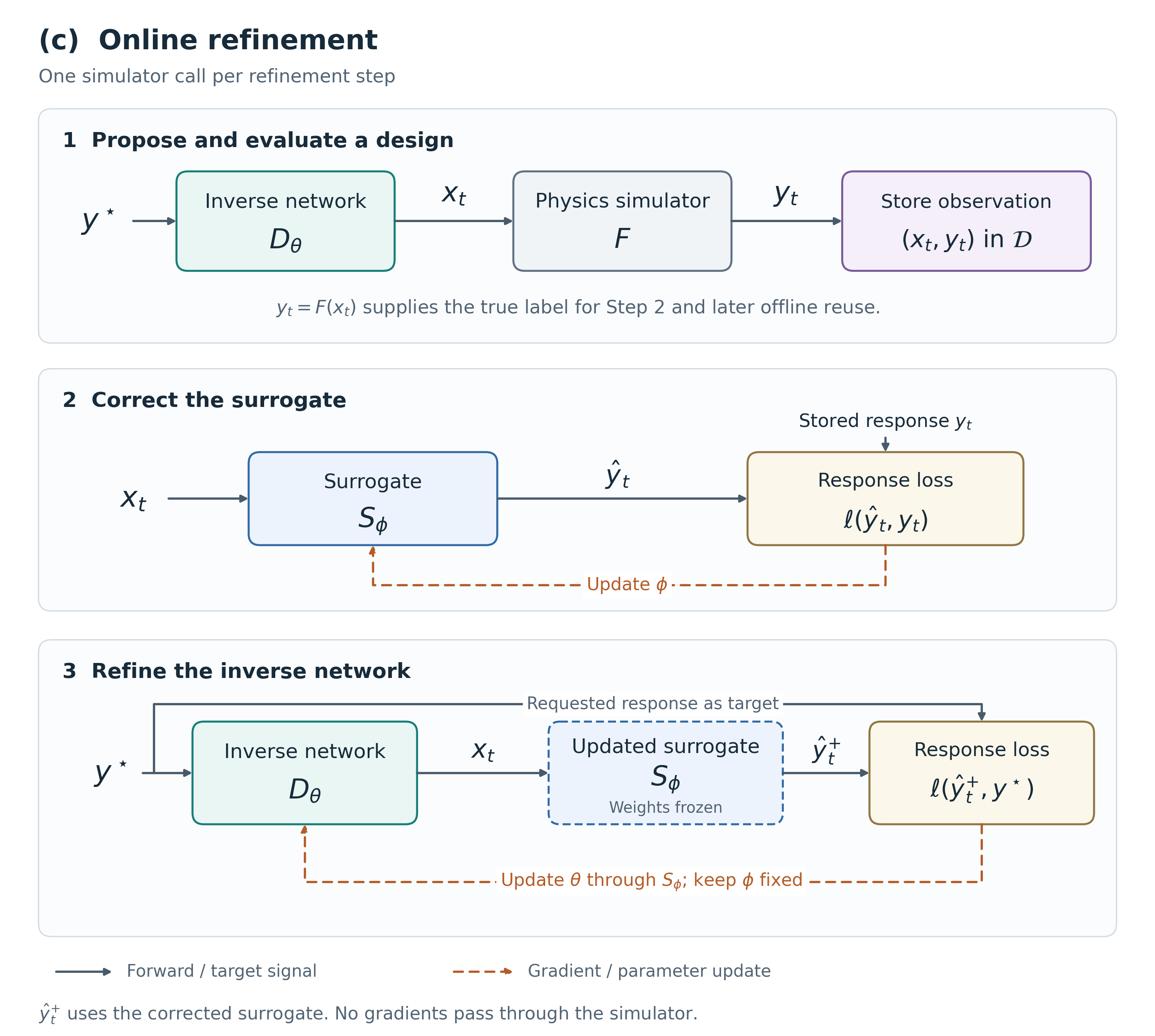}
  \caption{ORO online refinement.}
  \label{fig:flow_online}
  
\end{subfigure}
\hfill
\begin{subfigure}[b]{0.32\textwidth}
  \includegraphics[width=\linewidth]{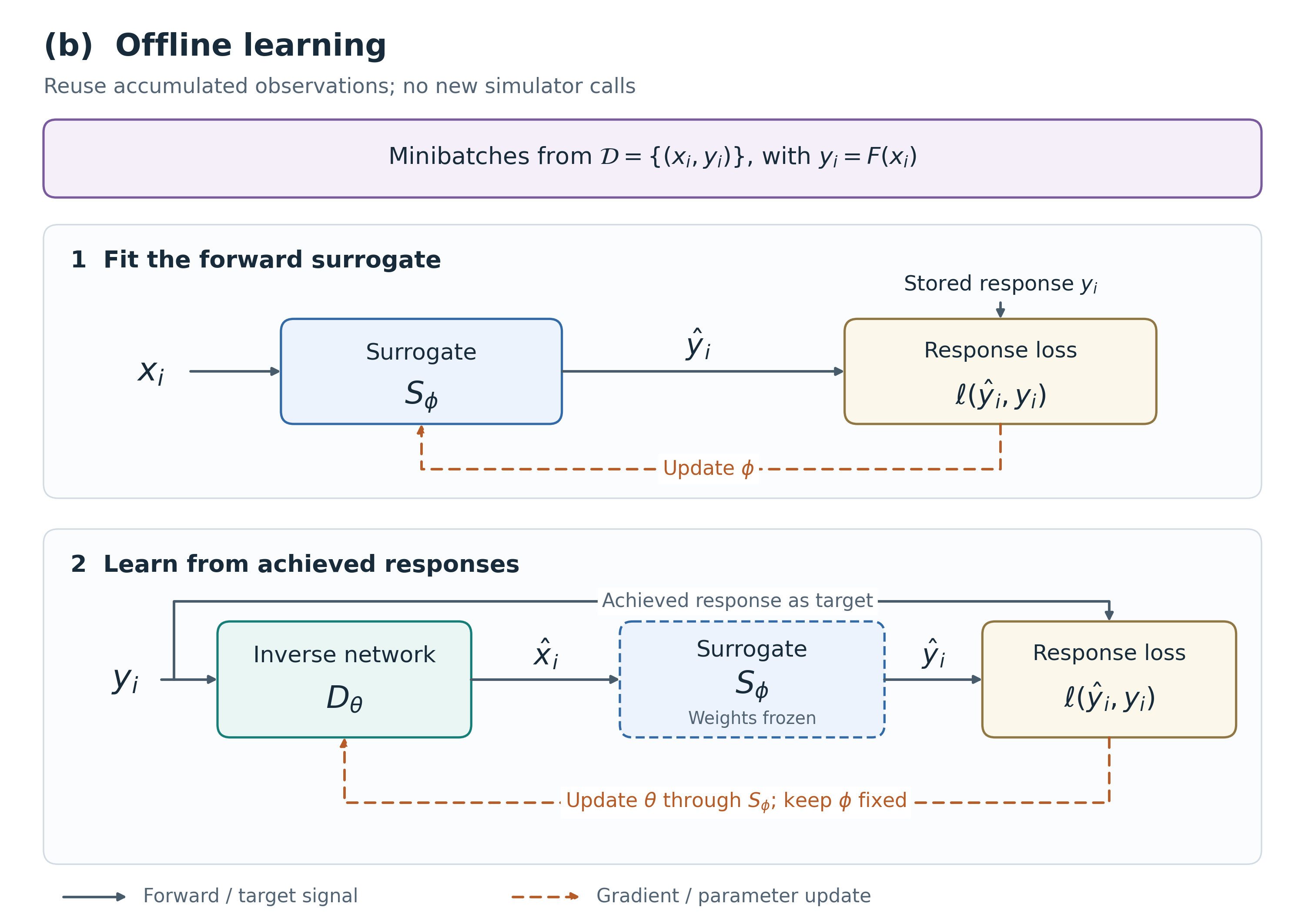}
  \caption{ORO offline learning}
  \label{fig:flow_offline}
\end{subfigure}
\caption{ORO alternates online refinement with offline reuse of accumulated design–response pairs. In online refinement (b), the inverse network is trained toward the requested response $y^*$, whose solution is unknown. In offline reuse (c), the surrogate is retrained on all of $\mathcal{D}$ and the inverse network is retrained through it, either toward achieved responses $y_i$ drawn from $\mathcal{D}$, each attainable by construction (as drawn in (c)), or toward the requested response $y^*$, as in our main experiments (Section~\ref{sec:offline}). In both phases the surrogate is fitted first and then frozen while the inverse network is fitted through it; gradients never pass through the simulator. Offline reuse consumes no simulator calls.}
\label{fig:algorithm}
\end{figure}

\subsection{Online refinement}
\label{sec:online}

Online refinement operates on the current target $y^{\star}$ and consumes one
simulator call per step (Fig.~\ref{fig:flow_online}). At step $t$ the design
network proposes
\begin{equation}
\hat{x}_{t}=D_{\theta}\!\left(y^{\star}\right),
\qquad
y_{t}=F(\hat{x}_{t}),
\label{eq:propose}
\end{equation}
where \eqref{eq:propose} contains the only evaluation of $F$ in the step. That
single observation is then used twice. First the surrogate is corrected at
exactly the point the design network currently proposes,
\begin{equation}
\mathcal{L}_{S}^{\text{on}}(\phi)=\ell\!\left(S_{\phi}(\hat{x}_{t}),\,y_{t}\right),
\label{eq:loss_s_online}
\end{equation}
so the correction lands where surrogate error most affects the next proposal.
Then, with $\phi$ held fixed, the design network is updated by differentiating
the target error through the freshly corrected surrogate,
\begin{equation}
\mathcal{L}_{D}^{\text{on}}(\theta)=
\ell\!\left(S_{\phi}\!\left(D_{\theta}(y^{\star})\right),\,y^{\star}\right).
\label{eq:loss_d_online}
\end{equation}
The simulator supplies ground-truth responses; the surrogate supplies the
approximate gradients that move the proposal. The pair $(\hat{x}_{t},y_{t})$ is
appended to $\mathcal{D}$ regardless of its error, and the best design observed
so far is retained as $x_{\text{best}}$.

\subsection{Offline reuse}
\label{sec:offline}

Every $N$ online steps, both networks are retrained on the whole accumulated
dataset $\mathcal{D}$ (Fig.~\ref{fig:flow_offline}). This phase makes \emph{no}
calls to $F$: it revisits observations already paid for. The surrogate is fitted
to the observed forward relationship,
\begin{equation}
\mathcal{L}_{S}^{\text{off}}(\phi)=
\frac{1}{|\mathcal{B}|}\sum_{(x_{i},y_{i})\in\mathcal{B}}
\ell\!\left(S_{\phi}(x_{i}),\,y_{i}\right),
\qquad \mathcal{B}\subset\mathcal{D},
\label{eq:loss_s_offline}
\end{equation}
and is then frozen while the design network is trained using each \emph{achieved}
response $y_{i}$ as its own input target,
\begin{equation}
\mathcal{L}_{D}^{\text{off}}(\theta)=
\frac{1}{|\mathcal{B}|}\sum_{(x_{i},y_{i})\in\mathcal{B}}
\ell\!\left(S_{\phi}\!\left(D_{\theta}(y_{i})\right),\,y_{i}\right).
\label{eq:loss_d_offline}
\end{equation}
Here the input to $D_{\theta}$ is a response $y_i$ the simulator has actually
produced, whose solution is known: the design $x_i$ that produced it. A design that
missed $y^{\star}$ is still a correct answer to the specification $y_i$, and because
the loss is in response space, $D_{\theta}$ may reproduce $y_i$ with any admissible
design rather than recover $x_i$, the appropriate supervision under non-uniqueness.
Reinterpreting achieved outcomes as goals relates to hindsight experience
replay~\citep{NIPS2017_453fadbd} and goal babbling~\citep{rolf2010goal}.

\paragraph{Requested-target alternative.} The offline phase can instead keep
training the inverse network on the problem actually being solved,
\begin{equation}
\mathcal{L}_{D}^{\text{off,req}}(\theta)=
\ell\!\left(S_{\phi}\!\left(D_{\theta}(y^{\star})\right),\,y^{\star}\right),
\label{eq:loss_d_offline_req}
\end{equation}
for the same number of updates, while the surrogate is still retrained on all of
$\mathcal{D}$. Objective~\eqref{eq:loss_d_offline} trains a general inverse map on
solved problems; \eqref{eq:loss_d_offline_req} exploits the retrained surrogate where
it matters and relies on it being accurate near a solution of $y^{\star}$. The two can
also alternate between offline rounds in a ratio $a{:}r$. Section~\ref{sec:mechanism}
compares them: the requested target is better on the resonance benchmark and equal on
the rational one, so the main experiments use~\eqref{eq:loss_d_offline_req} offline,
and~\eqref{eq:loss_d_offline} pre-trains $D_{\theta}$ on the warm-up set and
consolidates it between targets (Section~\ref{sec:method_sequential}). Every simulated
pair is used by the surrogate in both phases at the cost of a single call to $F$.

\subsection{Exploration}
\label{sec:explore}

Online refinement follows surrogate gradients and is therefore local. When sections
interact multiplicatively, as in the rational benchmark, a section at the wrong
frequency contributes little to the response and can only be relocated by a large move
of a few parameters through designs of higher error, which gradient steps do not make.
ORO therefore replaces scheduled proposals with exploratory ones: within each block of
$B_e$ refinement calls every $e$-th call is exploratory, and $e$ grows by one per
block. An exploratory call copies the best design found so far and resamples $k$
randomly chosen parameters uniformly within their bounds; the step otherwise proceeds
as usual, so the design is simulated and counted, corrects the surrogate, and enters
$\mathcal{D}$. With $e=2$, $B_e=2{,}000$ and $k=1$, $1{,}133$ of $2{,}400$ refinement
calls are exploratory. This is a mutation operator of the kind used in evolutionary
search, not an acquisition criterion: it involves no utility model and no inner
optimization.

\subsection{The full procedure}
\label{sec:algorithm}

Fig.~\ref{fig:flow_main} shows the complete cycle, and
Algorithm~\ref{alg:oro} in Appendix~\ref{app:algorithm} states it precisely. Starting from a warm-up set of
$n_{0}$ simulated designs, ORO alternates a burst of $N$ online steps with one
offline phase until the budget $B$ is exhausted, returning $x_{\text{best}}$.

\paragraph{Budget accounting.} Line~5 of Algorithm~\ref{alg:oro} is the only call to $F$ in the loop, so the
total simulator cost is exactly $n_{0}$ plus the number of online steps. The
offline phase contributes nothing, and $N$, $U$, $n_{b}$ and all learning rates
affect wall-clock but never simulator count. All comparisons in
Section~\ref{sec:experiments} equalise this total, including the warm-up $n_{0}$.
Exploratory calls are counted like any other.

\subsection{Sequences of targets}
\label{sec:method_sequential}

Nothing in Algorithm~\ref{alg:oro} ties $\mathcal{D}$, $\phi$ or $\theta$ to a
single specification. When several targets are to be designed, ORO
carries the dataset forward, so the warm-up is paid once. Between targets both
networks are re-initialised and pre-trained on the whole accumulated dataset, the
inverse network with the achieved-response objective~\eqref{eq:loss_d_offline};
during each target's run, the offline phase trains on the $N_{\text{sel}}$ designs
with the lowest error for the current target rather than on everything accumulated.
Both choices address known effects: networks warm-started from earlier training can
adapt worse than freshly initialised ones
\citep{ash2020warm,dohare2024loss}, and uniform replay of a large history dilutes the
data relevant to the current target. Because $D_{\theta}$ is conditioned on the
response, a pair collected for one target still informs the inverse map near any
later target that produces similar responses. Methods that search per target have no
equivalent mechanism: each run begins from nothing.
Section~\ref{sec:sequential} compares it with methods that solve each target
from scratch.
\section{Experiments}
\label{sec:experiments}

\subsection{Benchmark simulators}
\label{sec:benchmarks}

Evaluating an inverse-design method on a single physical problem confounds the
method with the problem. We therefore use three simulators chosen so that the
two principal difficulties of inverse design, non-uniqueness and
ill-conditioning, vary independently across them. All three map a design
$x\in\mathbb{R}^{15}$ ($K=5$ sections, three parameters each) to a response
sampled at $m=200$ points and share the same admissible box $\mathcal{X}$, so
they are interchangeable without altering any model, baseline, or
hyperparameter. Full definitions are given in Appendix~\ref{app:simulators}.

\textbf{Resonance} is a sum of $K$ Gaussian resonances; permuting the sections
leaves the response unchanged, so each response has $K!=120$ preimages.
\textbf{Rosenbrock} warps the normalised parameters into a curved Rosenbrock valley
centred on the target's own parameters and maps them through a fixed orthonormal basis;
the warp is invertible, so this simulator presents ill-conditioning without
non-uniqueness and serves as a control; its narrow curved valley is a standard test of
optimizers. \textbf{Rational} is a cascade of $K$
second-order sections, each a pole pair with a transmission zero, as in a filter
response; quality factors span $Q_i\in[3,100]$, so conditioning varies sharply near
high-$Q$ poles, and the $K!$ section symmetry is retained.

Table~\ref{tab:benchmarks} in Appendix~\ref{app:simulators} confirms that the
difficulties are present and separable. All three benchmarks are non-convex at every
sampled point, with median condition numbers of $10^3$ to $10^4$, reaching
$5.6\times10^{5}$. A single inverse network
trained with a parameter-space loss diverges on validation on the two simulators with
$K!$ equivalent solutions but not on the bijective control, because designs with
identical responses present the same input with different labels; supervision in
response space avoids this by construction (Appendix~\ref{app:direct}). Rosenbrock
serves as a control in the main comparison, where it isolates ill-conditioning; the
mechanism studies concentrate on the two non-unique benchmarks, which combine both
difficulties.

\subsection{Baselines and protocol}
\label{sec:protocol}

We compare ORO with five baselines covering the main alternatives for black-box
design. \textbf{GA} evaluates every candidate with $F$, standard practice in
electromagnetic design. \textbf{CMA-ES} \citep{hansen2001completely}, with
increasing-population restarts \citep{auger2005restart}, is a strong derivative-free
optimizer. \textbf{Bayesian optimization} (BO) uses a Gaussian process with an ARD
kernel and expected improvement, and \textbf{TuRBO} \citep{eriksson2019scalable}
restricts it to an adaptive trust region, which suits budgets of thousands of calls.
\textbf{AutoTandemML} \citep{GRBCIC2025113828} is the closest adaptive surrogate-based
inverse-design method: it collects a target-agnostic dataset by uncertainty sampling
and trains tandem networks on it. A genetic search over a surrogate trained once on a
fixed random dataset is omitted: it cannot correct its model for the current target,
and AutoTandemML represents surrogate-based design with adaptive data collection.
Implementation details are in Appendix~\ref{app:baselines}.

All methods search the same box $\mathcal{X}$, from which the targets are generated
(Appendix~\ref{app:simulators}); ORO's inverse network maps into it through bounded
output layers, and every proposal of every method lies in it. Every method receives
the same number of calls to $F$ per target, and all calls are counted, including
warm-up and dataset collection. ORO's warm-up is the first $400$ evaluations of the
GA run on the same target, so ORO and GA coincide up to call $400$. The direct-design comparison of
Section~\ref{sec:mechanism} and the sequential experiment of Section~\ref{sec:sequential}
use a task-agnostic Latin-hypercube warm-up instead. A single ORO
configuration is used for all benchmarks (Appendix~\ref{app:oro_hyper}). Performance
is best-so-far response error against calls, over $30$ targets drawn independently of
any training data. Because the error distribution is heavy-tailed, we report medians
and test differences with paired two-sided sign tests that exclude ties.

\subsection{Budget-matched comparison}
\label{sec:budget_matched}

\begin{figure}[t]
\centering
\begin{subfigure}[b]{0.32\textwidth}
  \includegraphics[width=\linewidth]{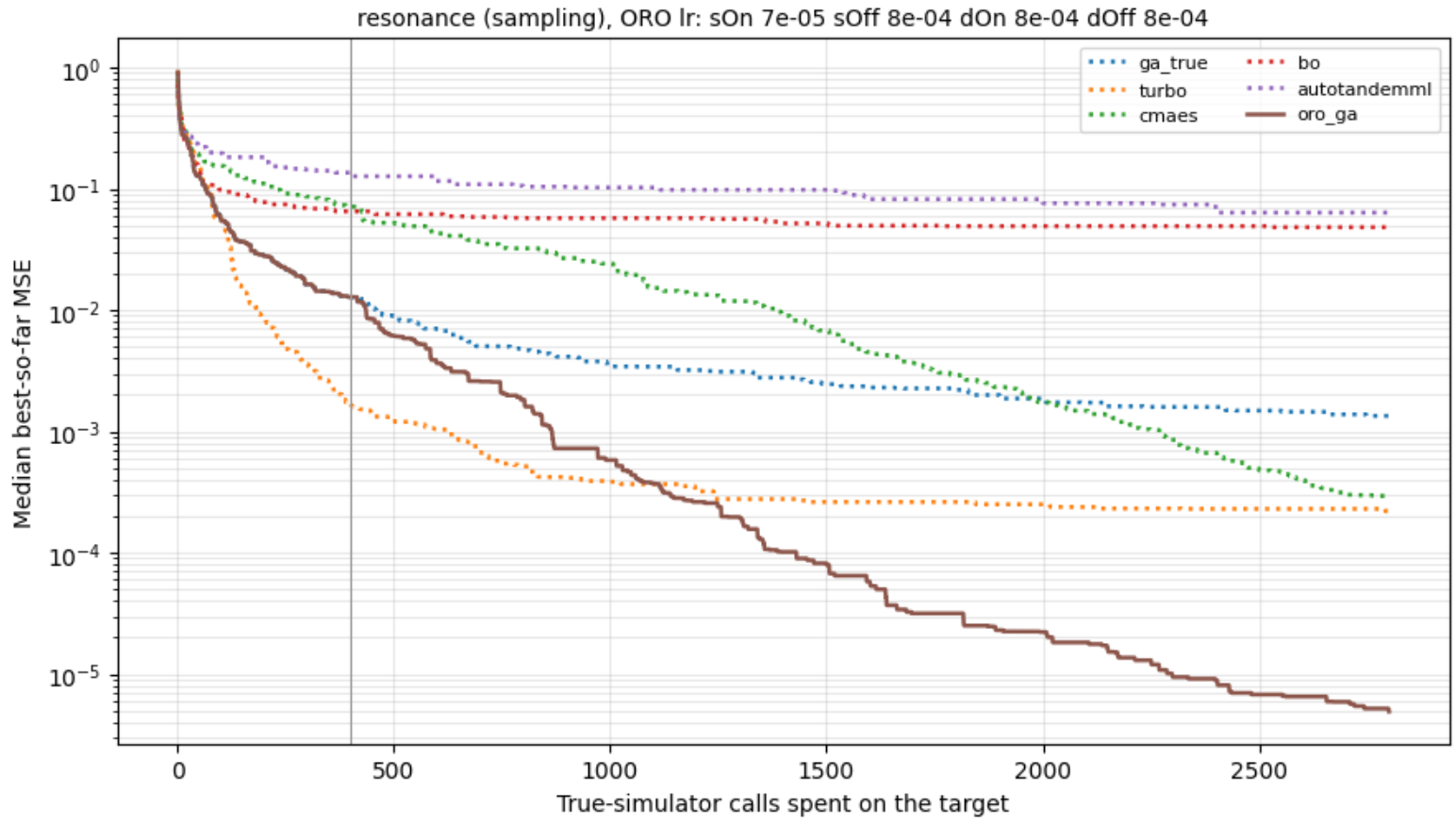}
  \caption{Resonance}
  \label{fig:budget_res}
\end{subfigure}
\hfill
\begin{subfigure}[b]{0.32\textwidth}
  \includegraphics[width=\linewidth]{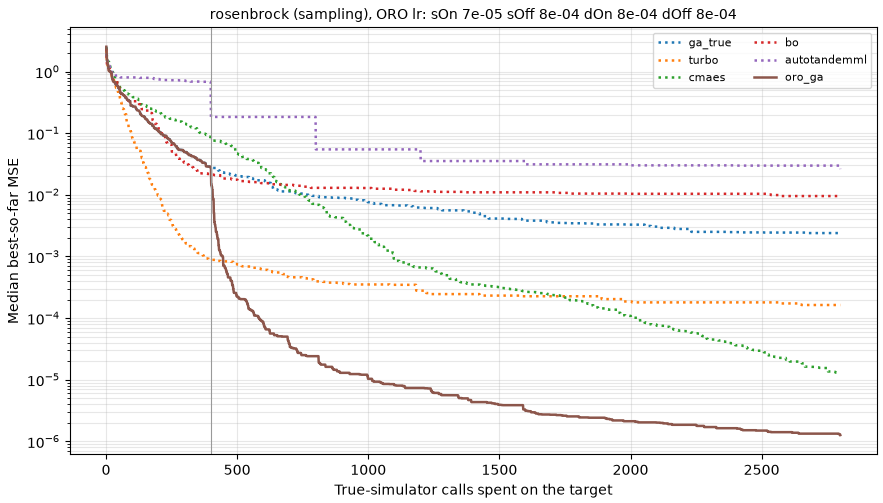}
  \caption{Rosenbrock}
  \label{fig:budget_ros}
\end{subfigure}
\hfill
\begin{subfigure}[b]{0.32\textwidth}
  \includegraphics[width=\linewidth]{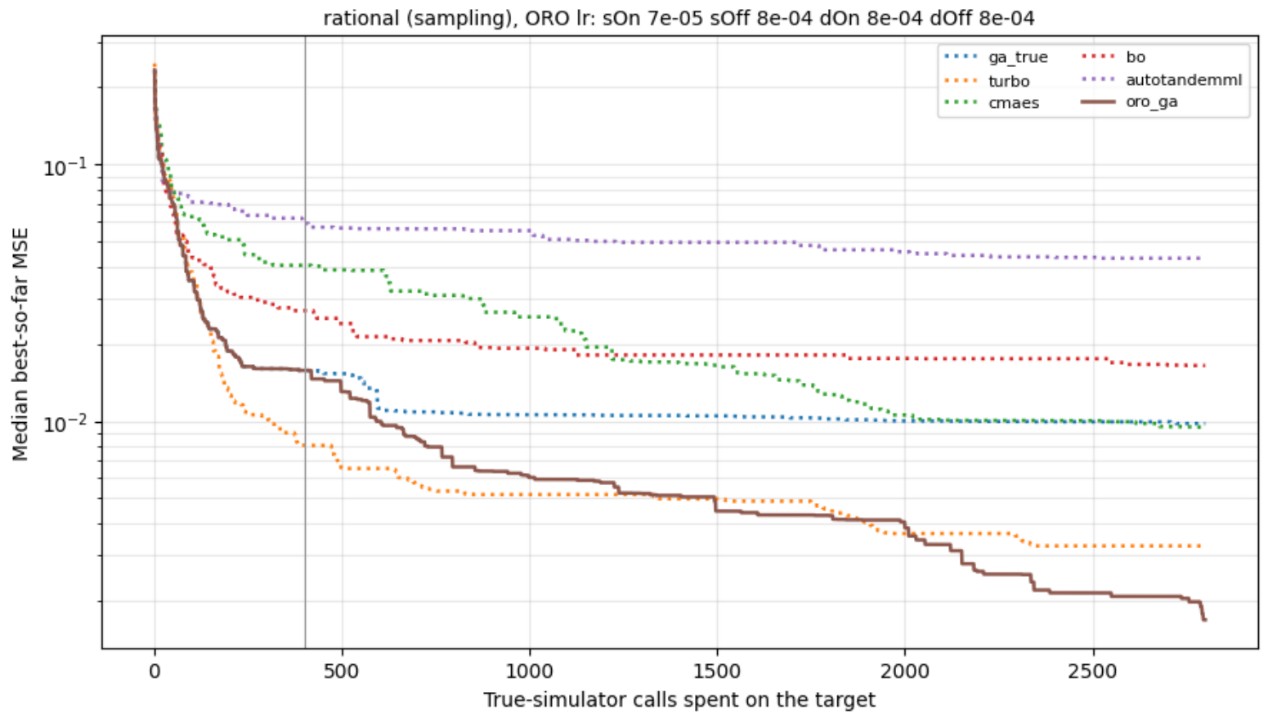}
  \caption{Rational}
  \label{fig:budget_rat}
\end{subfigure}
\caption{Median best-so-far response error over $30$ targets against
true-simulator calls. All methods receive identical budgets, every call is counted,
and all search the same bounds. ORO and GA share a warm-up search and coincide up to
call $400$. TuRBO leads at small budgets; ORO attains the lowest median at the full
budget.}
\label{fig:budget}
\end{figure}

\begin{table}[t]
\centering
\caption{Median response error over $30$ targets at $2{,}800$ calls per target,
and the number of targets on which ORO is better (paired two-sided sign test; ties
excluded).}
\label{tab:final}
\small
{
\resizebox{\linewidth}{!}{%
\begin{tabular}{lcccccc}
\toprule
 & \multicolumn{2}{c}{Resonance} & \multicolumn{2}{c}{Rosenbrock} & \multicolumn{2}{c}{Rational} \\
\cmidrule(lr){2-3}\cmidrule(lr){4-5}\cmidrule(lr){6-7}
Method & Median & ORO better & Median & ORO better & Median & ORO better \\
\midrule
ORO (proposed)   & $\mathbf{4.91{\times}10^{-6}}$ & -- & $\mathbf{1.26{\times}10^{-6}}$ & -- & $\mathbf{1.68{\times}10^{-3}}$ & -- \\
TuRBO            & $2.20{\times}10^{-4}$ & $26/30$ ($p{=}5.9{\times}10^{-5}$) & $1.63{\times}10^{-4}$ & $30/30$ ($p{=}1.9{\times}10^{-9}$) & $3.26{\times}10^{-3}$ & $16/30$ ($p{=}0.86$) \\
CMA-ES           & $2.95{\times}10^{-4}$ & $26/30$ ($p{=}5.9{\times}10^{-5}$) & $1.31{\times}10^{-5}$ & $25/30$ ($p{=}3.2{\times}10^{-4}$) & $9.48{\times}10^{-3}$ & $25/30$ ($p{=}3.2{\times}10^{-4}$) \\
GA               & $1.35{\times}10^{-3}$ & $28/30$ ($p{=}8.7{\times}10^{-7}$) & $2.40{\times}10^{-3}$ & $30/30$ ($p{=}1.9{\times}10^{-9}$) & $9.79{\times}10^{-3}$ & $29/30$ ($p{=}5.8{\times}10^{-8}$) \\
BO               & $4.83{\times}10^{-2}$ & $30/30$ ($p{=}1.9{\times}10^{-9}$) & $9.56{\times}10^{-3}$ & $30/30$ ($p{=}1.9{\times}10^{-9}$) & $1.65{\times}10^{-2}$ & $29/30$ ($p{=}5.8{\times}10^{-8}$) \\
AutoTandemML     & $5.92{\times}10^{-2}$ & $30/30$ ($p{=}1.9{\times}10^{-9}$) & $2.62{\times}10^{-2}$ & $30/30$ ($p{=}1.9{\times}10^{-9}$) & $4.32{\times}10^{-2}$ & $29/30$ ($p{=}5.8{\times}10^{-8}$) \\
\bottomrule
\end{tabular}}}
\end{table}

Figure~\ref{fig:budget} reports median best-so-far error against calls and
Table~\ref{tab:final} the error after $2{,}800$ calls. On resonance ORO attains a
median error of $4.9\times10^{-6}$, $45$ times below TuRBO and $60$ times below
CMA-ES, the two strongest baselines, and is better than every baseline on at least
$26$ of $30$ targets. On rational its median of $1.7\times10^{-3}$ is about half of
TuRBO's, but the paired comparison is even ($16/30$, $p=0.86$); ORO is better than
GA, CMA-ES, BO and AutoTandemML on $25$ to $29$ of $30$ targets.
On the Rosenbrock benchmark, which isolates ill-conditioning, ORO's lead is the most
consistent: its median of $1.3\times10^{-6}$ is the only one below $10^{-5}$, $10$ times
below CMA-ES, $129$ times below TuRBO and $1{,}900$ to $21{,}000$ times below GA, BO and
AutoTandemML, and it is better than CMA-ES on $25$ of $30$ targets and than every other
baseline on all $30$. This lead owes little to exploration: only $2.8$ exploratory calls
per target improve on the best design there (Section~\ref{sec:mechanism}).

The advantage depends on budget. TuRBO is the strongest method at small budgets on
all three benchmarks when, as here, each target starts from scratch. ORO, however, can
start from any existing dataset, even a random one unrelated to the target: in the
sequential setting of Section~\ref{sec:sequential} it leads TuRBO before spending a
single call on a new target. In the comparison here, ORO starts refining from the GA's first $400$ calls, whose median
error is then well above TuRBO's on every benchmark, and keeps improving until its
median overtakes TuRBO's, after roughly $500$ calls on Rosenbrock, $1{,}100$ on resonance
and $1{,}600$ on rational. Its advantage therefore does not come from its warm-up. Means are less favourable to ORO on the two non-unique benchmarks:
there TuRBO's mean final error is below ORO's
($3.4\times10^{-4}$ against $7.8\times10^{-4}$ on resonance, $3.8\times10^{-3}$ against
$4.8\times10^{-3}$ on rational), because ORO's error distribution has a heavier upper
tail: it reaches very low error on most targets and remains comparatively high on a few.
On Rosenbrock, with a single valley, ORO's mean equals its median and is the lowest. Appendix~\ref{app:examples} shows
the best designs of every method on representative targets.

\paragraph{Wall-clock.} Excluding warm-up and pre-training, ORO uses $336$ to
$425$\,s of GPU time per target across the benchmarks, against $98$ to $343$\,s for TuRBO
and $111$ to $401$\,s for BO; GA and CMA-ES need under a second
(Appendix~\ref{app:wallclock}). About $98\%$ of ORO's time is offline training,
independent of the simulator. These overheads matter only when simulations are cheap:
a CST simulation of our filter (Section~\ref{sec:filter}) takes about $5$ minutes on
average, so $2{,}800$ calls take
about $233$ hours, or $9.7$ days, per target, against $6$ to $7$ minutes of ORO's own
computation, under $0.1\%$ of the total. ORO's additional overhead over TuRBO, at most
$291$\,s, is less than one such call, so whenever ORO
reaches a given error in fewer calls it is also faster end to end. On resonance it
reaches TuRBO's final median error after about $1{,}250$ calls, saving some $1{,}550$
calls, or about $130$ hours of CST time per target.

\subsection{What drives ORO}
\label{sec:mechanism}
\label{sec:ablation}

Except for the offline-objective study, the analyses below use ORO without exploration and
with achieved-response training, so the effects they measure are those of learning
alone.

\paragraph{Offline retraining makes the surrogate's gradients useful.} Correcting
the surrogate at one design per step does not guarantee accurate local gradients.
Since our simulators are differentiable, we compare surrogate and true gradients at
regular probes, without using true gradients for optimization
(Appendix~\ref{app:gradients}). With offline retraining, the surrogate gradient is a
descent direction for the true error on $92\%$ of probes on resonance and $79\%$ on
rational, with median cosines of $0.77$ and $0.43$. Without the offline phase it is a
descent direction on only $47\%$ of probes on both, no better than chance, with
cosines near zero. This is why refinement stalls without offline reuse: correcting
the surrogate at single points fits its value at the proposal but not its slope.
Single-point online updates alone are therefore not a reliable learning signal; in ORO
they are small corrections, at a lower learning rate, between offline retrainings on
all data.
The online correction itself helps modestly on average, raising the share of descent
directions from $90\%$ to $92\%$ on resonance and from $77\%$ to $79\%$ on rational.
Section~\ref{sec:components} removes each component in turn: without offline surrogate
retraining the final median error is $442$ times higher on resonance and $116$ times on
Rosenbrock, and removing any other component raises it $6$ to $34$ times on those two
benchmarks.

\paragraph{What to train the inverse network on offline.} Table~\ref{tab:offline_target}
in Appendix~\ref{app:offline_target} compares the offline objectives of
Section~\ref{sec:offline}, with and without exploration. All variants perform the same
number of offline updates of both networks, so differences reflect the training target,
not more optimization steps or compute. On resonance, training on the requested target lowers the final error
by a per-target median factor of $2.9$ without exploration and $7.7$ with it, and is
better on $23$ and $27$ of $30$ targets; alternating rounds is as good or better. With
exploration every ratio from $1{:}2$ to $1{:}6$ beats achieved-response training, and
$1{:}4$ and $1{:}6$ do not differ significantly from each other. On rational the
choice matters little: per-target ratios lie between $0.89$ and $0.99$, and the only
significant difference is a $4\%$ gain from $1{:}1$ alternation without exploration
($25/30$). We read this as a question of how far the retrained surrogate can be
trusted near the solution: the requested objective exploits it there, which pays off
when it is accurate, as on resonance, while the achieved-response objective is safer
when it is not. The benefit of the offline phase therefore comes mainly from
retraining the surrogate and continuing to train the inverse network through it, not
from achieved-response targets specifically. Choosing the ratio adaptively from the
surrogate's prediction error is left to future work.

\paragraph{Exploration.} On average $14.9$ exploratory calls per target on resonance
and $44.1$ on rational improve on every earlier design in the comparison of
Section~\ref{sec:budget_matched}, so ORO is best described as a learned refiner combined
with mutation of its best design; on Rosenbrock, whose single valley needs no
relocation, only $2.8$ do, so there ORO's progress comes mainly from its own proposals. Mutation alone does not explain the results: in the
offline-objective runs on resonance most improvements come from ORO's own proposals
($29$ to $36$ per target, against $14$ to $17$ from exploration;
Appendix~\ref{app:offline_target}), and GA, a mutation-based search, loses to ORO on at
least $28$ of $30$ targets on every benchmark. Exploration matters most on rational: the median error of every offline objective
plateaus near $1.0\times10^{-2}$ without it, and reaches $2.8$--$4.6\times10^{-3}$ with
it. On resonance its effect is modest: under achieved-response training, exploration
lowers the per-target error by about $10\%$ in median, on $18$ of $30$ targets, which is
not significant ($p=0.36$). This matches the failure mode that motivated it
(Section~\ref{sec:explore}).

\paragraph{The inverse network versus direct design optimization.} Keeping ORO's
surrogate, schedule and updates but optimizing a single design vector through the
corrected surrogate, an online neural adjoint \citep{NEURIPS2020_007ff380}, is worse:
from a Latin-hypercube warm-up of $400$ designs (Appendix~\ref{app:direct_design}), the
inverse network is better on $28$ of $30$ resonance targets after $1{,}000$ calls
(median $2.1\times10^{-3}$ against $1.0\times10^{-2}$, $p=8.7\times10^{-7}$), and ahead in
median but not significantly on rational ($1.0$ against $1.3\times10^{-2}$; $18/30$,
$p=0.36$). The inverse network thus contributes beyond the corrected surrogate.

\subsection{Designing many targets with few simulations}
\label{sec:sequential}

We test whether ORO's accumulated data make many targets cheap to design without
the large up-front dataset of surrogate-based design. Thirty
targets are presented in sequence with $1{,}000$ new calls each. ORO pays for a
$400$-design Latin-hypercube warm-up once and carries its dataset forward as in
Section~\ref{sec:method_sequential}; every baseline solves each target from scratch
with the same $1{,}000$ calls. A simpler scheme, carrying both networks forward and
replaying all data uniformly, gave no benefit on the rational benchmark in our earlier
experiments and failed to match a GA searching a $30{,}000$-sample offline surrogate on
$8$ of $60$ instances, against $3$ of $60$ when resetting; the scheme of Section~\ref{sec:method_sequential} addresses the two likely
causes, loss of plasticity and dilution of relevant data.

\begin{figure}[t]
\centering
\begin{subfigure}[b]{0.32\textwidth}
  \includegraphics[width=\linewidth]{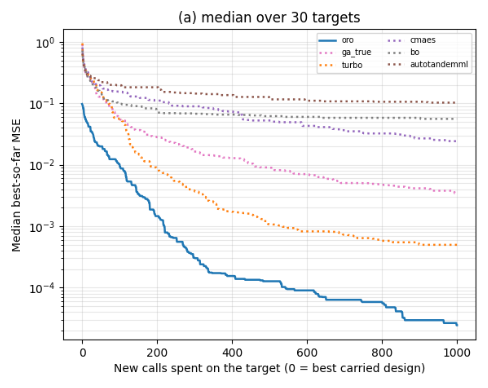}
  \caption{Resonance}
\end{subfigure}
\hfill
\begin{subfigure}[b]{0.32\textwidth}
  \includegraphics[width=\linewidth]{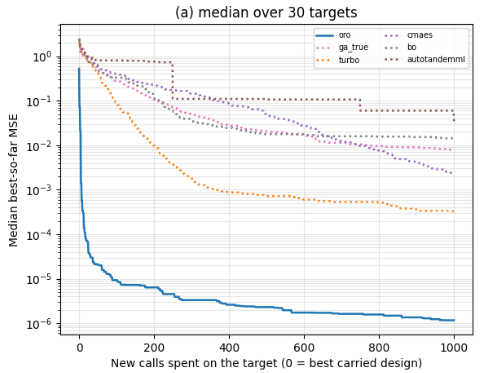}
  \caption{Rosenbrock}
\end{subfigure}
\hfill
\begin{subfigure}[b]{0.32\textwidth}
  \includegraphics[width=\linewidth]{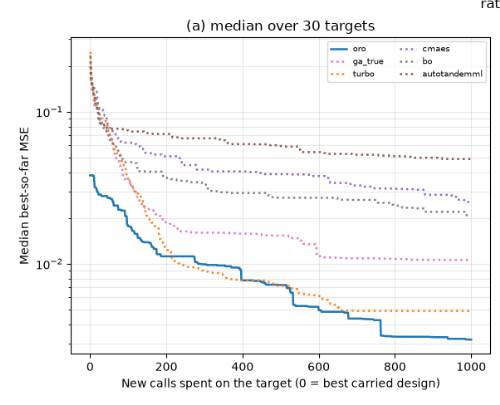}
  \caption{Rational}
\end{subfigure}
\caption{Sequential design: median best-so-far error over $30$ targets against new
calls on each target. For ORO, call $0$ is the best design carried from earlier targets
and the warm-up; the baselines start each target from scratch.}
\label{fig:sequential}
\end{figure}

On resonance, ORO's median error after $1{,}000$ calls per target is $2.4\times10^{-5}$,
$20$ times below TuRBO's; it is better than every baseline on at least $28$ of $30$
targets, has the lowest mean error, and leads TuRBO at every budget
(Figure~\ref{fig:sequential}; Appendix~\ref{app:sequential}). Later targets also start from lower error: the rank
correlation between target index and error is $-0.46$ ($p=0.011$) at the start of each
target and $-0.54$ ($p=0.0019$) after $300$ calls, although it is not significant at
$50$, $100$ or $1{,}000$ calls. On rational, ORO's median of $3.2\times10^{-3}$ is again
the lowest, against $4.9\times10^{-3}$ for TuRBO and $1.1\times10^{-2}$ to
$4.9\times10^{-2}$ for the others; it beats GA on $26$ and CMA-ES, BO and AutoTandemML on
all $30$ targets, but, as in Section~\ref{sec:budget_matched}, not TuRBO significantly
($18/30$, $p=0.36$), and no trend over the sequence is significant. On Rosenbrock,
ORO's median of $1.2\times10^{-6}$ is $282$ times below TuRBO's and $2{,}100$ to
$28{,}000$ times below the other baselines; it is better than every baseline on all $30$
targets, and slightly below its own budget-matched median at $2{,}800$ calls. It is also
the one benchmark on which later targets reach lower error at every budget: the rank
correlation between target index and error lies between $-0.51$ and $-0.87$
($p\le0.0042$) at $0$, $50$, $100$, $300$ and $1{,}000$ calls. Its head start is small (median carried error about $0.5$), yet its median passes
TuRBO's final median within $50$ calls. On resonance and
rational the carried data give ORO a head start, paid for by its warm-up and the simulations of
earlier targets: with no calls yet spent on a new target, its best carried design has a
median error of about $10^{-1}$ on resonance and $4\times10^{-2}$ on rational, and on
rational ORO stays ahead of TuRBO for about the first hundred calls. The baselines
start each target from scratch, as is standard for per-target optimizers; seeding them
with the carried data is possible and would narrow this head start. These results use a
task-agnostic Latin-hypercube warm-up, not a target-directed GA search.

\subsection{Band-pass filter design with a full-wave solver}
\label{sec:filter}

\begin{figure}[t]
\centering
\begin{subfigure}[b]{0.32\textwidth}
  \includegraphics[width=0.8\linewidth]{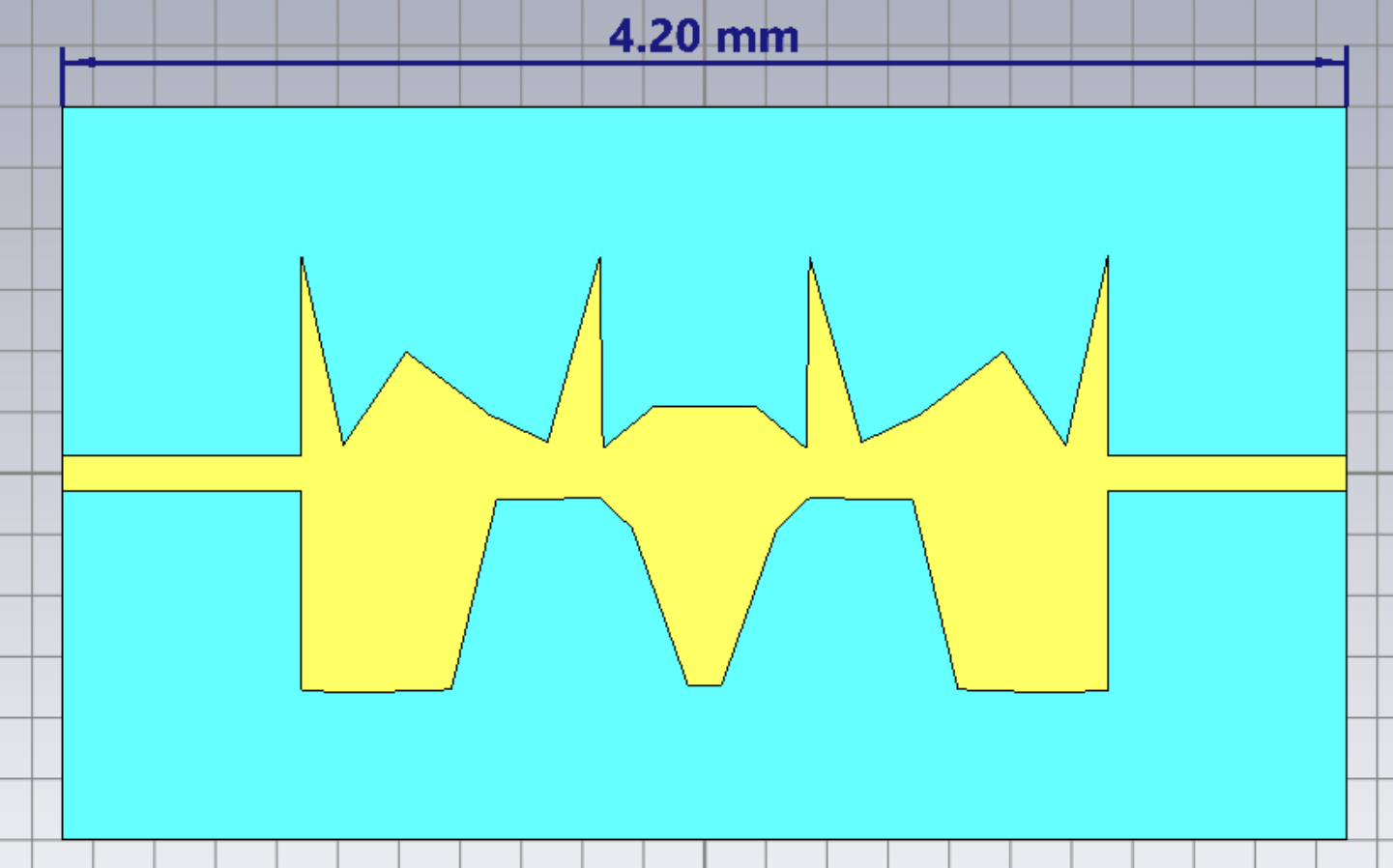}
  \caption{Geometry produced by ORO}
  \label{fig:filter_design}
\end{subfigure}
\hfill
\begin{subfigure}[b]{0.32\textwidth}
  \includegraphics[width=0.9\linewidth]{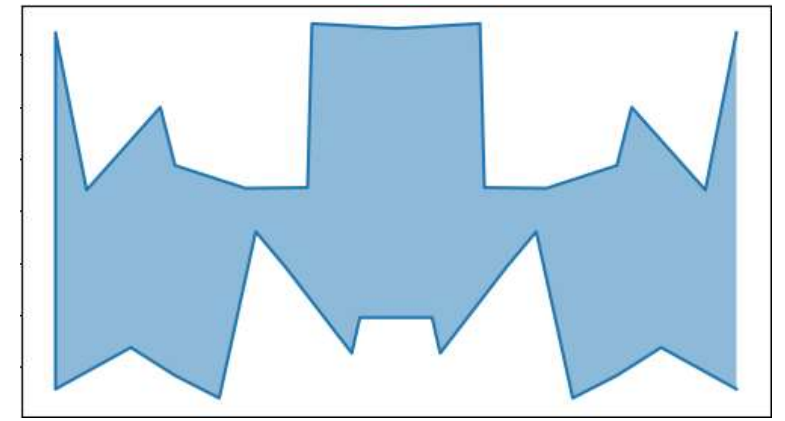}
  \caption{Reference geometry}
  \label{fig:filter_target}
\end{subfigure}
\hfill
\begin{subfigure}[b]{0.32\textwidth}
  \includegraphics[width=\linewidth]{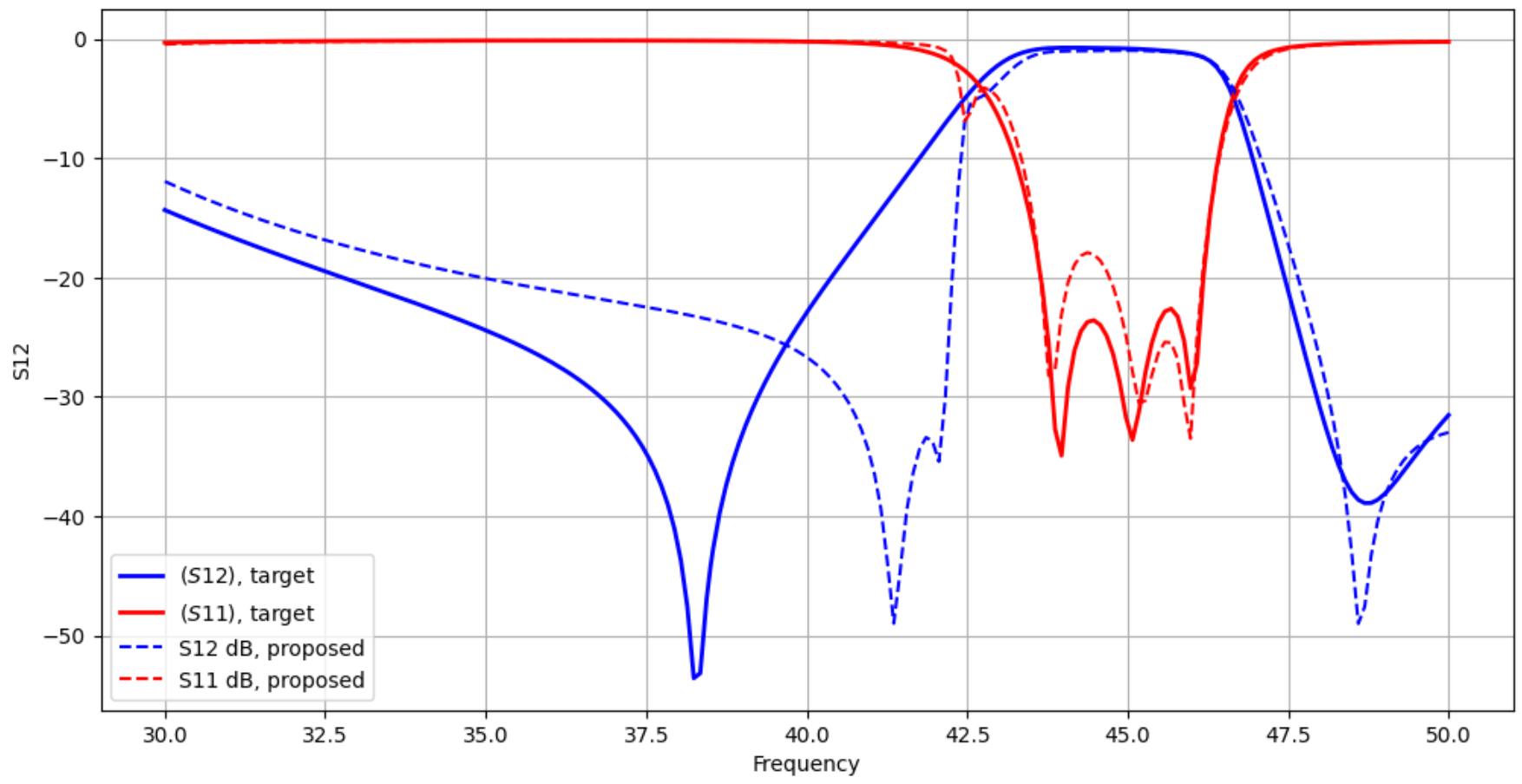}
  \caption{Simulated responses}
  \label{fig:filter_results}
\end{subfigure}
\caption{Band-pass filter design with a commercial full-wave solver. (a) Geometry
produced by ORO. (b) Geometry of \citet{elbadrawy2026vector}, whose simulated
response defines the target. (c) Simulated $S_{11}$ and $S_{21}$ of both designs.
ORO meets the specification in approximately $1{,}200$ simulations against
$16{,}300$, and sharpens the upper band edge through an additional transmission
zero.}
\label{fig:filter_fig}
\end{figure}

To test whether the benchmark conclusions transfer to production, we apply ORO to
millimetre-wave band-pass filter design in CST, adopting unchanged the geometry
and parameterisation of \citet{elbadrawy2026vector}: a single-layer irregular
vector-graphics structure defined by controllable vertices under a symmetry
constraint, with $S_{11}$ and $S_{21}$ simulated over $30$--$50$\,GHz. That work
generates its dataset in advance by Latin hypercube sampling and searches a
convolutional surrogate with a genetic algorithm, using $16{,}300$ full-wave
simulations. ORO instead
queries the solver only along the trajectory its inverse network induces toward
the requested response, and reaches a compliant design in approximately
$1{,}200$ simulations with the same geometry and solver settings.
Here the target is the simulated response of the reference design, so a design
counts as compliant when it reproduces that response, which is known to be attainable,
and meets the same passband specification (Figure~\ref{fig:filter_fig}); the second
experiment below designs from explicit numerical limits instead. At about $5$ minutes
per simulation, ORO's $1{,}200$ simulations take about
$100$ hours, against about $1{,}360$ hours ($57$ days) for $16{,}300$.
We did not time ORO's network training in this run; on the synthetic
benchmarks it takes under $10$ minutes per $2{,}800$ calls, so even with the filter's
larger networks it is a small fraction of the simulation time.

Both designs meet the passband specification, and ORO's geometry adds a second
transmission zero in the upper stopband, which was not requested: the response-space
loss admits any geometry that reproduces the target.

\paragraph{Design from a specification.} A filter is usually specified by limits
rather than by a reference curve, so we also design directly from a specification
mask with no reference design behind it: $|S_{21}|\geq-1$\,dB and $|S_{11}|\leq-20$\,dB
over the passband ($42.5$--$47.5$\,GHz), and $|S_{21}|<-30$\,dB and $|S_{11}|\geq-0.1$\,dB
in the stopbands (below $42$ and above $48$\,GHz). The response loss is replaced by a
smooth penalty on constraint violations (Appendix~\ref{app:filter}), so the target
need not be attainable. After $1{,}620$ simulations, the best design meets the
insertion-loss limit over most of the passband and the stopband limit over most of the
stopbands, but not everywhere, so the specification is met only in part
(Appendix~\ref{app:filter}). Over four independent runs, ORO reached designs of similar
quality after $1{,}000$ to $1{,}700$ simulations, every run attaining a comparably small
specification loss.

\subsection{Component ablation}
\label{sec:components}

\begin{figure}[t]
\centering
\begin{subfigure}[b]{0.32\textwidth}
  \includegraphics[width=\linewidth]{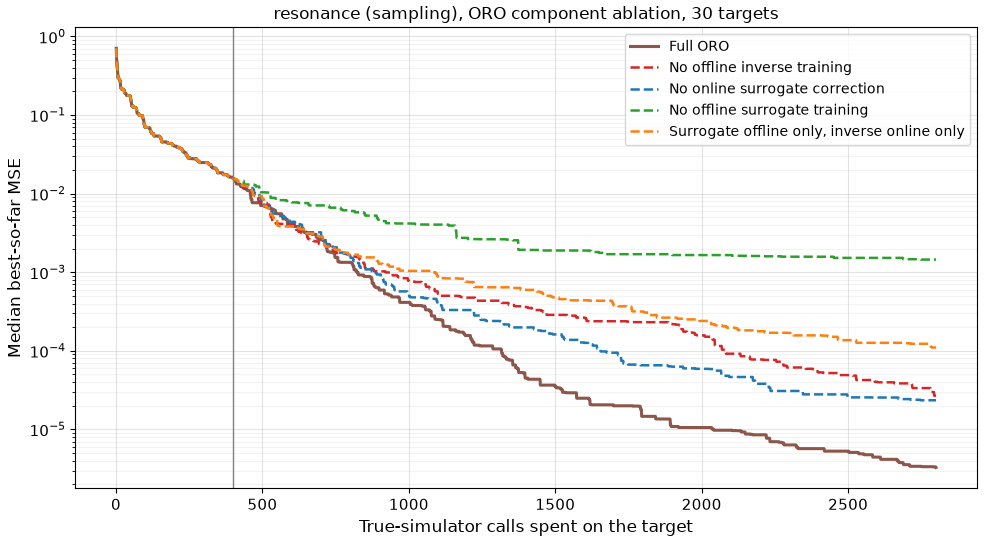}
  \caption{Resonance}
  \label{fig:components_res}
\end{subfigure}
\hfill
\begin{subfigure}[b]{0.32\textwidth}
  \includegraphics[width=\linewidth]{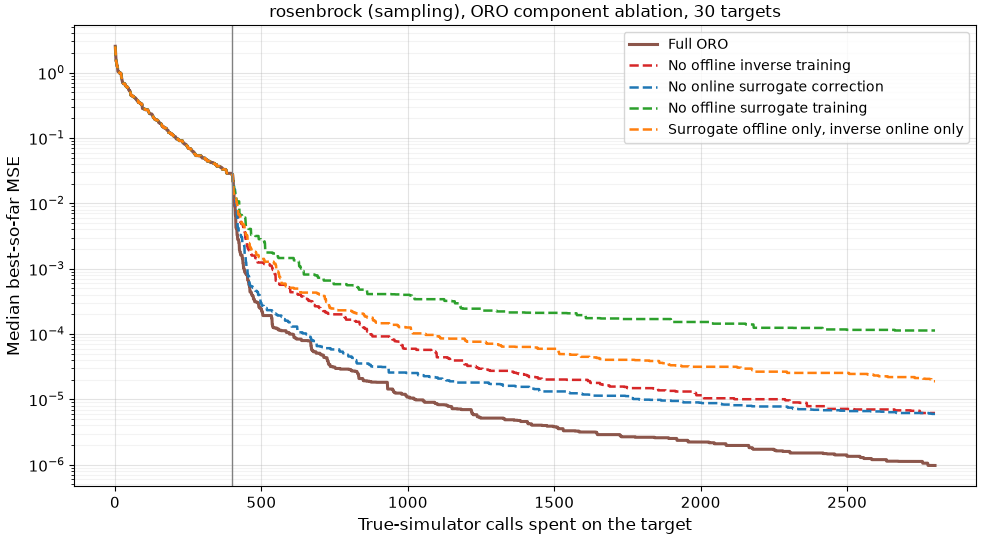}
  \caption{Rosenbrock}
  \label{fig:components_ros}
\end{subfigure}
\hfill
\begin{subfigure}[b]{0.32\textwidth}
  \includegraphics[width=\linewidth]{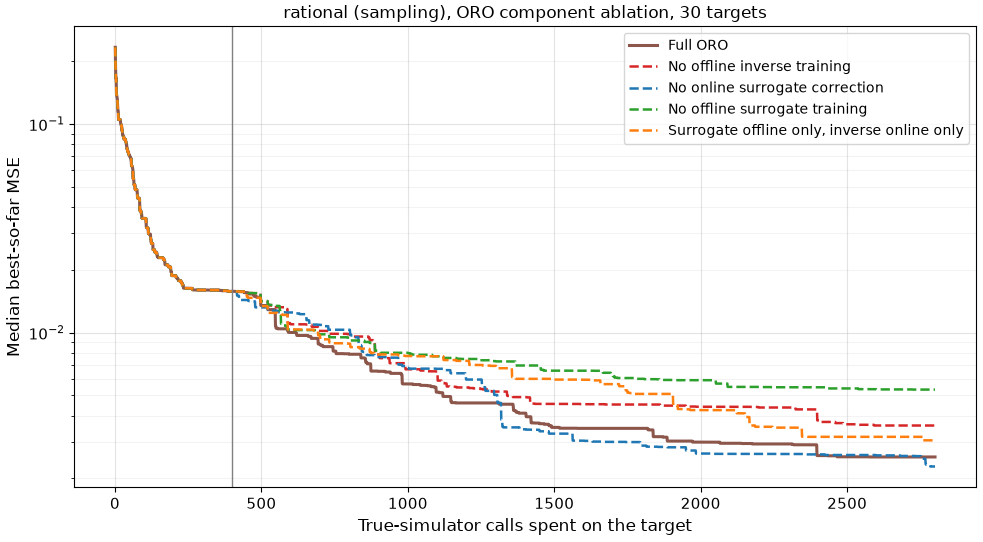}
  \caption{Rational}
  \label{fig:components_rat}
\end{subfigure}
\caption{Component ablation: median best-so-far error over $30$ targets against
true-simulator calls. Full ORO is solid; each dashed curve removes one component. All
variants share the warm-up and coincide up to call $400$.}
\label{fig:components}
\end{figure}

Retraining the surrogate offline is the component ORO depends on most, and on resonance
and Rosenbrock every other component also helps. Each variant removes one part of ORO
and keeps everything else: the inverse network is not trained offline; the surrogate is
not corrected online; the surrogate is not retrained offline; or the surrogate is trained
only offline and the inverse network only online, which removes the first two together.
The runs follow the protocol of Section~\ref{sec:budget_matched}: $30$ targets,
$2{,}800$ calls, GA's $400$-call warm-up, requested-target offline training and an
exploratory jump on every second call. Each variant shares its warm-up, pre-trained
networks and random seeds with full ORO, so the variants differ only in the part
removed. Full ORO is a separate run here, so its medians differ slightly from
Table~\ref{tab:final}. Table~\ref{tab:components} gives the factors and
Figure~\ref{fig:components} the median curves.

\begin{table}[h]
\caption{Component ablation: each variant's median error at $2{,}800$ calls divided by
full ORO's, over $30$ targets. In parentheses, the number of targets on which full ORO
is better (paired sign test; $p\le0.016$ except without online surrogate correction on
rational, $p=0.58$).}
\label{tab:components}
\centering
\small
\begin{tabular}{lccc}
\toprule
 & Resonance & Rosenbrock & Rational \\
\midrule
Full ORO (median error) & $3.3{\times}10^{-6}$ & $9.8{\times}10^{-7}$ & $2.5{\times}10^{-3}$ \\
\midrule
No offline inverse training & $8.2\times$ ($28/30$) & $6.4\times$ ($30/30$) & $1.4\times$ ($22/30$) \\
No online surrogate correction & $7.2\times$ ($29/30$) & $6.2\times$ ($30/30$) & $0.90\times$ ($17/30$) \\
No offline surrogate retraining & $442\times$ ($30/30$) & $116\times$ ($30/30$) & $2.1\times$ ($27/30$) \\
Surrogate offline only, inverse online only & $34\times$ ($30/30$) & $19\times$ ($30/30$) & $1.2\times$ ($24/30$) \\
\bottomrule
\end{tabular}
\end{table}

Without offline surrogate retraining, the median error at $2{,}800$ calls is $442$ times
higher on resonance, $116$ times on Rosenbrock and $2.1$ times on rational, and full ORO
is better on at least $27$ of $30$ targets on each benchmark. Its curves flatten soon
after the warm-up, and on resonance its own proposals almost stop improving on the best
design: $2.2$ new bests per target against $45$ for full ORO, so $87\%$ of its error
reduction comes from the exploratory jumps. This matches the gradient diagnostics of
Section~\ref{sec:mechanism}, where a surrogate corrected only at single points gives
gradients no better than chance.

Online surrogate correction and offline inverse training each matter on resonance and
Rosenbrock: removing either raises the median error $6$ to $8$ times, and full ORO is
better on at least $28$ of $30$ targets. Removing both costs more than either alone, $34$
times on resonance and $19$ on Rosenbrock. On rational every effect is smaller: $1.4$
times without offline inverse training ($22/30$) and $1.2$ times with the surrogate
trained only offline ($24/30$).

On rational, removing the online correction makes no significant difference at the full
budget ($0.90$ times, $17/30$, $p=0.58$), but the curves show that it helps earlier:
full ORO's median is lower over most of the first $1{,}300$ calls, after which the two
are close, the variant without it slightly lower (Figure~\ref{fig:components_rat}). As
the buffer grows, offline retraining alone keeps the surrogate accurate, and further
single-point corrections add little. This suggests an online surrogate learning rate
that decreases with the number of calls, so that the surrogate stands in for the
simulator more as it becomes reliable; it would cost no additional simulator calls. On
resonance and Rosenbrock the online correction helps throughout the run, so the
decrease would have to be gradual there. We leave this to future work.

\section{Conclusion}
\label{sec:conclusion}

We introduced ORO, which designs with an expensive black-box simulator by using a
target-conditioned inverse network as both design model and query rule, correcting a
surrogate online, retraining it offline at no extra simulation cost, and exploring from
its best design on a fixed schedule. With every call counted and identical bounds, it
attains the lowest median error at the full budget against five baselines, $129$ and
$45$ times below TuRBO on the ill-conditioned Rosenbrock and resonance benchmarks, and
its behaviour traces to offline retraining, which makes the surrogate's gradients reduce
the true error. Across a sequence of targets it keeps the lowest median error on
all three benchmarks, by factors of $282$ and $20$ over TuRBO on Rosenbrock and
resonance. On a full-wave filter it reaches a compliant design in about
$1{,}200$ simulations.

\subsection*{AI use statement}

In this work, we used generative AI tools for polishing the text, discovering related works, and assisting with code generation. We have not used generative AI tools for generating core research ideas, primary claims, or experimental results, and the rest of the required disclosure tasks are not applicable to this work. Additionally, we used generative AI tools for basic syntax formatting and proofreading. We have reviewed all AI-assisted work. Specifically, all LLM-generated code was thoroughly revised, tested, and validated for correctness by the authors. AI-suggested related literature was manually verified against primary sources, and polished writing was reviewed to ensure scientific accuracy and preservation of the original intent. We take responsibility for the final content of this work, including text, claims or artifacts produced with the aid of generative AI.

\bibliography{Ref}
\bibliographystyle{iclr2027_conference}

\appendix

\section{Benchmark simulators}
\label{app:simulators}

All three simulators map a design $x\in\mathbb{R}^{15}$ to a response sampled at
$m=200$ uniformly spaced points $f\in[0,1]$. The design is organised as $K=5$
sections with three parameters each, $x=(a_{1:K},\,c_{1:K},\,w_{1:K})$. Targets
and warm-up designs are drawn uniformly from the box
\begin{equation}
a_i\in[0.5,\,1.5],\qquad c_i\in[0.1,\,0.9],\qquad w_i\in[0.03,\,0.12],
\label{eq:app_box}
\end{equation}
and, where a simulator requires it, parameters are normalised to the unit cube,
$u=(x-x_{\min})/(x_{\max}-x_{\min})$, using the bounds of~\eqref{eq:app_box}.
The same box and parameter layout are shared by all three, so every model,
baseline and hyperparameter is used unchanged across them.

\paragraph{Resonance.} A sum of $K$ Gaussian resonances with amplitudes $a_i$,
centres $c_i$ and widths $w_i$,
\begin{equation}
R(f)=\sum_{i=1}^{K} a_i\exp\!\left(-\frac{(f-c_i)^2}{2w_i^2}\right).
\label{eq:app_resonance}
\end{equation}
The sum is invariant to permutation of the section index, so any response has
$K!=120$ parameter preimages that differ only in labelling.

\paragraph{Rosenbrock.} The normalised parameters are passed through the
curved-valley warp
\begin{equation}
z_0=u_0,\qquad z_i=10\left(u_i-u_{i-1}^{2}\right),\quad i=1,\dots,14,
\label{eq:app_warp}
\end{equation}
and mapped to a response by $R=Bz$, where $B\in\mathbb{R}^{200\times15}$ has
orthonormal columns, obtained once from the QR decomposition of a fixed Gaussian
matrix and held constant throughout. Because $B^{\top}B=I$,
\begin{equation}
\ell\!\left(R(x),R(x^{\star})\right)=\tfrac{1}{m}\left\|z(u)-z(u^{\star})\right\|_2^{2},
\label{eq:app_rosen}
\end{equation}
so the response-error landscape searched by every method is exactly a Rosenbrock
valley centred on the target's own parameters, with value zero at $x=x^{\star}$
(verified numerically to within $2.4\times10^{-7}$). The warp is recursively
invertible, $u_i=z_i/10+u_{i-1}^{2}$, and $B$ has full column rank, so the
simulator is a bijection onto its range: it presents ill-conditioning without
non-uniqueness.

\paragraph{Rational.} A cascade of $K$ second-order sections, each with a complex
pole pair and a transmission zero. The three parameter groups map, through their
normalised values $(u_{a},u_{c},u_{w})$, to a quality factor, a pole frequency
and a zero frequency,
\begin{equation}
Q_i=Q_{\min}\!\left(\tfrac{Q_{\max}}{Q_{\min}}\right)^{u_{a,i}},\qquad
f_{p,i}=0.1+0.8\,u_{c,i},\qquad
f_{z,i}=f_{p,i}+0.25\left(u_{w,i}-\tfrac12\right),
\label{eq:app_rational_map}
\end{equation}
with $Q_{\min}=3$, $Q_{\max}=100$, and $f_{z,i}$ clipped to $[0.02,\,1.2]$. The
response is
\begin{equation}
H(f)=\prod_{i=1}^{K}\frac{f^{2}-f_{z,i}^{2}}{\left(f^{2}-f_{p,i}^{2}\right)+\jmath\,f\,f_{p,i}/Q_i},
\qquad
R(f)=\frac{|H(f)|^{2}}{1+|H(f)|^{2}}.
\label{eq:app_rational}
\end{equation}
The product is evaluated as a sum of logarithms and clamped before
exponentiation, so high-$Q$ poles cannot overflow. The bounded output mimics
$|S_{11}|^{2}$ and lies in $[0,1]$. Over $500$ random designs the response spans
the full interval with no degenerate curves, and only $4\%$ and $9\%$ of samples
saturate near $1$ and $0$ respectively. Like the resonance simulator, the product
is invariant to permutation of the section index.

\paragraph{Conditioning measurement.} For each simulator a single target
$y^{\star}=F(x^{\star})$ is fixed, and at $25$ further designs drawn
from~\eqref{eq:app_box} we compute the Hessian of
$\ell\!\left(F(x),y^{\star}\right)$ with respect to $x$ by automatic
differentiation. A point is counted as non-convex if the smallest eigenvalue is
below $-10^{-8}$, and the condition number is the ratio of largest to smallest
absolute eigenvalue. These are the quantities reported in
Table~\ref{tab:benchmarks}. Automatic differentiation is used only for this
characterisation; no method under comparison has access to simulator gradients.

\begin{table}[h]
\centering
\caption{Structural properties of the three benchmark simulators. Divergence is
the ratio of final to minimum validation loss when a single inverse network is
trained with a parameter-space loss (Appendix~\ref{app:direct}); values near $1$
indicate no divergence. Response error is the mean true-simulator error of the
designs produced on held-out targets.}
\label{tab:benchmarks}
\small
\begin{tabular}{lccc}
\toprule
 & Resonance & Rosenbrock & Rational \\
\midrule
Global optima              & $K!=120$ & $1$ & $K!=120$ \\
Non-convex (of 25 pts)     & 25 & 25 & 25 \\
Hessian cond., median      & $1.4\times10^{3}$ & $3.4\times10^{3}$ & $2.9\times10^{4}$ \\
Hessian cond., max         & $5.8\times10^{4}$ & $1.6\times10^{4}$ & $5.6\times10^{5}$ \\
\midrule
Direct-learning divergence & $1.71\times$ & $1.11\times$ & $1.56\times$ \\
Response error, direct     & $0.603$ & $0.0277$ & $0.144$ \\
Response error, tandem     & $0.311$ & $0.0083$ & $0.102$ \\
\bottomrule
\end{tabular}
\end{table}

\section{Direct inverse learning}
\label{app:direct}

This appendix gives the full experiment summarised in
Section~\ref{sec:benchmarks}. A single inverse network $D_{\theta}$, with the
architecture used throughout the paper, is trained to regress design parameters
from responses with the parameter-space loss
\begin{equation}
\mathcal{L}_{\text{direct}}(\theta)=\mathrm{MSE}\!\left(x,\,D_{\theta}(y)\right).
\label{eq:app_direct}
\end{equation}
For each simulator we draw $8{,}000$ training and $1{,}500$ validation designs
from~\eqref{eq:app_box}, standardise the responses using training-set
statistics, and train for $420$ epochs with Adam at learning rate $10^{-3}$ and
batch size $128$. Architecture, data size, optimizer and schedule are identical
across the three simulators, so they differ only in the structure of the
mapping. As a control on the same data we also train the tandem construction of
\citet{liu2018training}: a forward surrogate is fitted first, then frozen while an
inverse network is trained through it with the response-space loss
$\mathrm{MSE}\!\left(y,\,S_{\phi}(D_{\theta}(y))\right)$. Both are finally scored
by the true-simulator response error of the designs they produce on the
validation targets, which is the quantity that matters for design.

\begin{figure}[h]
\centering
\includegraphics[width=\linewidth]{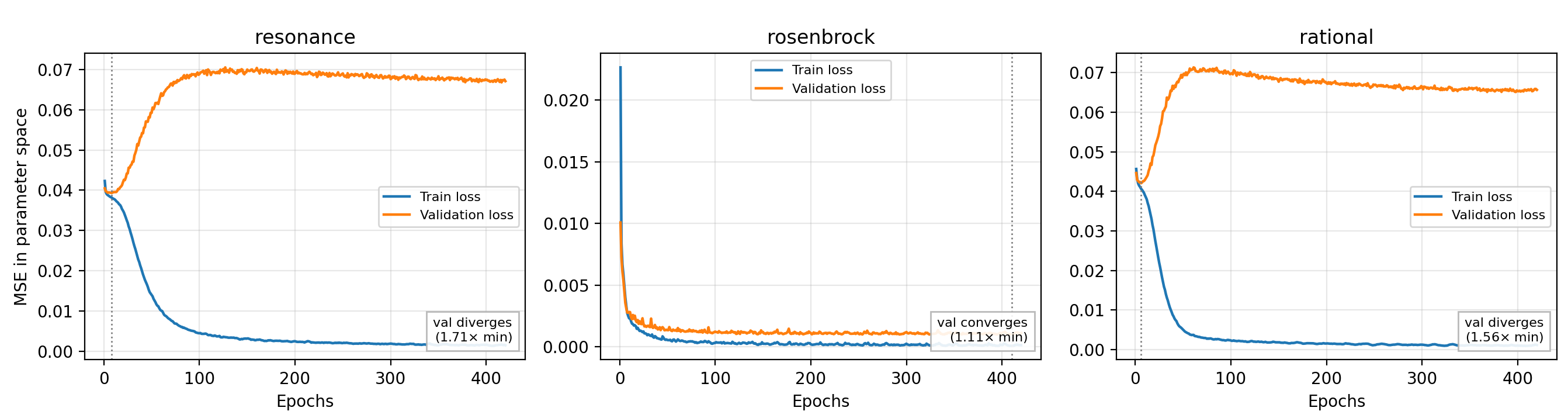}
\caption{Training and validation loss of a single inverse network trained with
the parameter-space loss~\eqref{eq:app_direct} on $8{,}000$ designs. Validation
loss reaches its minimum within the first ten epochs and then rises on the two
simulators with $K!$ equivalent solutions; it does not on the bijective
Rosenbrock control. The dotted line marks the epoch of minimum validation loss.}
\label{fig:direct}
\end{figure}

Figure~\ref{fig:direct} shows the result. On resonance and rational, validation
loss is minimised at epochs $7$ and $5$ respectively and then rises steadily,
ending $1.71\times$ and $1.56\times$ above that minimum while training loss falls
by more than an order of magnitude. On Rosenbrock the minimum occurs at epoch
$409$ and the final value is within $1.11\times$ of it, with validation tracking
training throughout.

The difference is structural. On a simulator with $K$ interchangeable sections,
two designs $x_1\neq x_2$ with $F(x_1)=F(x_2)$ present the network with the same
input and different labels. The minimiser of~\eqref{eq:app_direct} over such a
pair is their average, which in general realises neither response and need not
be a valid design at all. The network fits these conflicting labels by
memorisation on the training set, which does not transfer. On the bijective
control no such conflict exists, and the same network generalises.

Table~\ref{tab:app_direct} confirms that this is not a small-sample effect.
Doubling the training set increases the divergence on both non-unique simulators,
from $1.55\times$ to $1.71\times$ on resonance and from $1.43\times$ to
$1.56\times$ on rational, while the bijective control is essentially unchanged.
This is the expected direction: a larger sample contains more pairs of distinct
designs with near-identical responses, so the conflicting supervision becomes
more prevalent rather than less.

\begin{table}[h]
\centering
\caption{Direct parameter-space learning versus the tandem response-space
construction, at two training-set sizes. Divergence is the ratio of final to
minimum validation loss under~\eqref{eq:app_direct}. Response error is the mean
true-simulator error of the designs produced on the validation targets.}
\label{tab:app_direct}
\small
\begin{tabular}{llccc}
\toprule
Training designs & Quantity & Resonance & Rosenbrock & Rational \\
\midrule
\multirow{3}{*}{$4{,}000$}
 & Divergence, direct      & $1.55\times$ & $1.09\times$ & $1.43\times$ \\
 & Response error, direct  & $0.641$ & $0.0392$ & $0.151$ \\
 & Response error, tandem  & $0.320$ & $0.0087$ & $0.163$ \\
\midrule
\multirow{3}{*}{$8{,}000$}
 & Divergence, direct      & $1.71\times$ & $1.11\times$ & $1.56\times$ \\
 & Response error, direct  & $0.603$ & $0.0277$ & $0.144$ \\
 & Response error, tandem  & $0.311$ & $0.0083$ & $0.102$ \\
\bottomrule
\end{tabular}
\end{table}

In response space the tandem construction improves on direct learning by
$1.9\times$, $3.3\times$ and $1.4\times$ at $8{,}000$ designs. At $4{,}000$
designs it is marginally worse than direct learning on the rational simulator
($0.163$ against $0.151$) and better elsewhere. The tandem network can only be as
good as the surrogate it optimises through, and the rational simulator is the
hardest to fit, having the most severe local conditioning in
Table~\ref{tab:benchmarks}; with more data the surrogate improves and the tandem
advantage appears. This dependence on surrogate quality is one motivation for
correcting the surrogate online, where the inverse network is actually
proposing, rather than fitting it once to a fixed sample.

\section{Implementation details}
\label{app:hyper}

\subsection{Networks for the benchmark simulators}
\label{app:arch}

Both networks are fully connected with ReLU activations. The forward surrogate
$S_{\phi}$ maps the $15$ design parameters to the $200$-point response through
hidden layers of width $128$, $256$ and $256$. The inverse design network
$D_{\theta}$ maps the $200$-point response to the $15$ design parameters through
four hidden layers of width $256$, followed by a linear head.

Admissibility is enforced in the output layer of $D_{\theta}$ rather than by
constrained optimization. Each output passes through a logistic sigmoid scaled to its
interval of the box~\eqref{eq:app_box},
\begin{equation}
x_j = x_{\min,j}+\left(x_{\max,j}-x_{\min,j}\right)\sigma(h_j),
\label{eq:app_heads}
\end{equation}
where $h$ denotes the pre-activation output and $\sigma$ the logistic sigmoid. The
admissible set is therefore the box from which targets and warm-up designs are drawn,
and it is the box searched by every baseline. An earlier version of this work gave the
inverse network a wider output range than some baselines searched; all experiments
reported here, including the component ablation and the gradient probes, use the common
box.

\subsection{Algorithm}
\label{app:algorithm}

\begin{algorithm}[h]
\caption{ORO: Online Refinement, Offline Reuse}
\label{alg:oro}
\begin{algorithmic}[1]
\Require target $y^{\star}$; simulator $F$; budget $B$; online burst $N$;
offline updates $U$; batch size $n_{b}$
\State $\mathcal{D}\leftarrow\{(x_{i},F(x_{i}))\}_{i=1}^{n_{0}}$;\quad
       $b\leftarrow n_{0}$
       \Comment{warm-up, charged to $B$}
\State pre-train $S_{\phi}$ with \eqref{eq:loss_s_offline}, then $D_{\theta}$
       with \eqref{eq:loss_d_offline}
\While{$b<B$}
  \For{$t=1$ \textbf{to} $\min(N,\,B-b)$}
      \Comment{\textbf{online}}
    \State $\hat{x}\leftarrow D_{\theta}(y^{\star})$ (or an exploratory move
           from $x_{\text{best}}$ on scheduled steps, Section~\ref{sec:explore});\quad
           $y\leftarrow F(\hat{x})$;\quad $b\leftarrow b+1$
      \Comment{only call to $F$}
    \State update $\phi$ with \eqref{eq:loss_s_online}
    \State freeze $\phi$; update $\theta$ with \eqref{eq:loss_d_online}
    \State $\mathcal{D}\leftarrow\mathcal{D}\cup\{(\hat{x},y)\}$;\ update
           $x_{\text{best}}$
  \EndFor
  \For{$u=1$ \textbf{to} $U$}
      \Comment{\textbf{offline}, $0$ calls to $F$}
    \State sample $\mathcal{B}\sim\mathcal{D}$, $|\mathcal{B}|=n_{b}$;\ update
           $\phi$ with \eqref{eq:loss_s_offline}
  \EndFor
  \State freeze $\phi$
  \For{$u=1$ \textbf{to} $U$}
    \State sample $\mathcal{B}\sim\mathcal{D}$, $|\mathcal{B}|=n_{b}$;\ update
           $\theta$ with \eqref{eq:loss_d_offline} or \eqref{eq:loss_d_offline_req}
  \EndFor
\EndWhile
\State \Return $x_{\text{best}}$
\end{algorithmic}
\end{algorithm}

\subsection{ORO hyperparameters}
\label{app:oro_hyper}

A single configuration is used for all benchmarks in the budget-matched comparison;
the sequential experiment differs only where Table~\ref{tab:app_hyper} states. All
networks are trained with Adam, and each network uses one optimizer shared across the
online and offline phases, with a learning rate per phase
(Table~\ref{tab:app_hyper}). Offline minibatches have size $\min(128,|\mathcal{D}|)$.

\begin{table}[h]
\centering
\caption{ORO hyperparameters for the budget-matched comparison
(Section~\ref{sec:budget_matched}), the offline-objective study
(Section~\ref{sec:mechanism}) and the sequential experiment
(Section~\ref{sec:sequential}).}
\label{tab:app_hyper}
\footnotesize
{
\resizebox{\linewidth}{!}{%
\begin{tabular}{lccc}
\toprule
 & Budget-matched & Offline objective & Sequential \\
\midrule
Simulator calls per target            & $2{,}800$ & $2{,}000$ & $1{,}000$ new \\
Warm-up $n_0$                         & $400$ (GA, per target) & $400$ (GA, per target) & $400$ (LHS, once) \\
Pre-training epochs                   & $2{,}500$ & $2{,}500$ & $2{,}500$ \\
Learning rate, $S$ online / offline   & $7{\times}10^{-5}$ / $8{\times}10^{-4}$ & same & same \\
Learning rate, $D$ online / offline   & $8{\times}10^{-4}$ / $8{\times}10^{-4}$ & same & same \\
Learning-rate decay                   & none & none & none \\
Online burst $N$                      & $3$ & $3$ & $3$ \\
Offline training per phase, $S$ and $D$ & $15$ epochs & $15$ epochs & $15$ epochs \\
Offline objective for $D$             & requested~\eqref{eq:loss_d_offline_req} & varied & requested \\
Exploration                           & $e{=}2$, $B_e{=}2{,}000$, $k{=}1$ & $e{=}2$, $B_e{=}2{,}000$, $k{=}1$, or none & $e{=}2$ within each target \\
Buffer $N_{\text{sel}}$               & -- & -- & $2{,}500$ (res., Ros.), $800$ (rat.) \\
Consolidation between targets         & -- & -- & $3{,}000$ epochs, fresh weights \\
\bottomrule
\end{tabular}}}
\end{table}

\paragraph{Offline compute.} Each offline phase trains each network for $15$
epochs over its training set, so $U=15\lceil|\mathcal{D}|/n_b\rceil$ in
Algorithm~\ref{alg:oro}. Its cost therefore grows with the dataset, from $400$ to
$2{,}800$ designs over a budget-matched run, and the wall-clock times of
Appendix~\ref{app:wallclock} include it. In the sequential experiment the training set
is bounded by the buffer $N_{\text{sel}}$ plus the target's own designs, so the cost per
phase stays bounded as the history grows.

\paragraph{Optimizer state.} Each network uses a single optimizer shared across
both phases, so momentum and step counts are continuous across phase switches.
Two optimizers over the same parameters would maintain moment estimates that go
stale against each other's updates.

Each phase has its own learning rate; the online surrogate rate is the smallest,
so that a single-point correction refines rather than overwrites what the offline phase
learned from the whole dataset.

In the sequential experiment, each offline phase of a target trains on the
$N_{\text{sel}}$ designs of the whole accumulated dataset with the lowest error for that
target, together with the target's own new designs. The buffer size $N_{\text{sel}}$ is
the only setting chosen per landscape, in preliminary runs: a small, target-specific
buffer ($800$) suits the rational benchmark, whose sharp features need a locally
accurate surrogate, while a larger one ($2{,}500$) helps on the smoother resonance
benchmark and is also used on Rosenbrock.

\subsection{Baselines}
\label{app:baselines}

\paragraph{GA.} Population $20$, elite fraction $0.2$, uniform crossover, and Gaussian
mutation applied per parameter with probability $0.2$ and standard deviation $0.08$ of
that parameter's range, with offspring clipped to the box. Every call simulates a new
design: elites are carried without re-simulation, and the run stops exactly at the
budget. A single GA run per target supplies ORO's warm-up, its first $400$
evaluations, so the two coincide over that range by construction.

\paragraph{CMA-ES.} CMA-ES \citep{hansen2001completely} as implemented in
\texttt{pycma}, in coordinates normalised to the unit box, with initial step size $0.3$
and increasing-population restarts \citep{auger2005restart}: the population doubles at
each restart. Candidates are clipped to the box and the run stops exactly at the
budget.

\paragraph{Bayesian optimization.} A Gaussian process with a squared-exponential kernel
and automatic relevance determination (one lengthscale per design dimension),
initialised with $20$ uniformly random designs. Observed errors are standardised to
zero mean and unit variance before every fit, since a GP has a zero-mean prior whereas
response errors are strictly positive. Once more than $300$ points have been observed,
the GP is fitted to a random subset of $300$ that always includes the incumbent; this
bounds the cubic cost of exact GP inference over thousands of evaluations, and a
random rather than best-only subset retains information about poor regions. TuRBO,
below, addresses the same issue by modelling locally and is the stronger BO baseline.
The kernel hyperparameters are refitted every $25$ evaluations by maximising the
marginal likelihood ($60$ Adam steps, learning rate $0.05$, log-lengthscales clipped to
$[-3,3]$). Expected improvement is maximised by scoring $3{,}000$ random candidates and
refining the best $8$ with $40$ steps of gradient ascent.

\paragraph{TuRBO.} TuRBO-1 \citep{eriksson2019scalable}, starting from $20$
scrambled Sobol designs and proposing batches of $8$ by Thompson sampling over $3{,}000$
candidates in the trust region, with the trust-region schedule of the original method.
The GP has a Mat\'ern-$5/2$ ARD kernel and models the logarithm of the response error,
which is positive and heavy-tailed; it is fitted to the $400$ evaluated designs nearest
the trust-region centre, since TuRBO's model is local by design, while every
evaluation is counted. When a GP fit fails, the last successful hyperparameters are
reused: this occurred in $6$ of $10{,}000$ fits on resonance, $30$ of $10{,}025$ on
Rosenbrock and none of $9{,}955$ on rational.

\paragraph{AutoTandemML.} Following \citet{GRBCIC2025113828}: $20$ Latin-hypercube
designs, then batches of $5$ designs, each maximising the predictive uncertainty of a
$150$-tree random forest by particle-swarm search ($10$ particles, $10$ iterations),
with tandem networks trained on the collected data. It collects one target-agnostic
dataset per benchmark. At each checkpoint its inverse network, trained on all data
collected so far, proposes a design for every target, and that proposal is simulated;
every call is counted, both data collection and checkpoint evaluations, and its
best-so-far error also includes collected designs that happen to lie near the target.
We made one adaptation. With an unbounded inverse output, its proposals left the box
on every target, so we bound the output by a sigmoid scaled to the box, as for ORO,
and report this version, which performed better in a preliminary comparison.
Both tandem networks are fully connected with ReLU activations and hidden layers of
width $64$, $128$, $256$, $128$ and $64$, on min-max-scaled responses and designs. The
forward network is trained first with a root-mean-square-error loss, then frozen while
the inverse network is trained through it with the same loss in response space; each
uses Adam (learning rate $10^{-3}$, batch size $32$) for up to $2{,}000$ epochs, with
early stopping (patience $10$) on a $10\%$ validation split, keeping the best weights.
The networks are retrained on the collected data at every checkpoint, every $400$
calls.

\subsection{Band-pass filter design}
\label{app:filter}

The filter geometry is defined by the planar coordinates of $15$ vertices on one
side of the structure's symmetry axis, giving $30$ design parameters. Each vertex
is confined to its own interval along the propagation axis, so that vertex order
is preserved, and to a fixed transverse range on its side of the axis; the
network output is mapped to these bounds from the unit interval. The target and
simulated responses are the magnitudes of $S_{11}$ and $S_{21}$ in decibels over
$30$--$50$\,GHz, concatenated.

Both networks are fully connected with ReLU activations and hidden layers of
width $256$, $512$ and $512$. The inverse network ends in a sigmoid, so every
proposal lies in the admissible box by construction. Refinement uses learning
rate $10^{-4}$ for both networks, an online burst of $N=3$, and $20$ offline
epochs per network per cycle with minibatch size $128$.
The target of this run is the simulated response of the reference design
(Section~\ref{sec:filter}). Unlike the synthetic experiments, it trains the inverse
network offline with the achieved-response objective~\eqref{eq:loss_d_offline}, and
instead of the scheduled exploration of Section~\ref{sec:explore} it adds random noise
to some proposals at the start of the run, then stops.

\begin{figure}[h]
\centering
\includegraphics[width=0.9\linewidth]{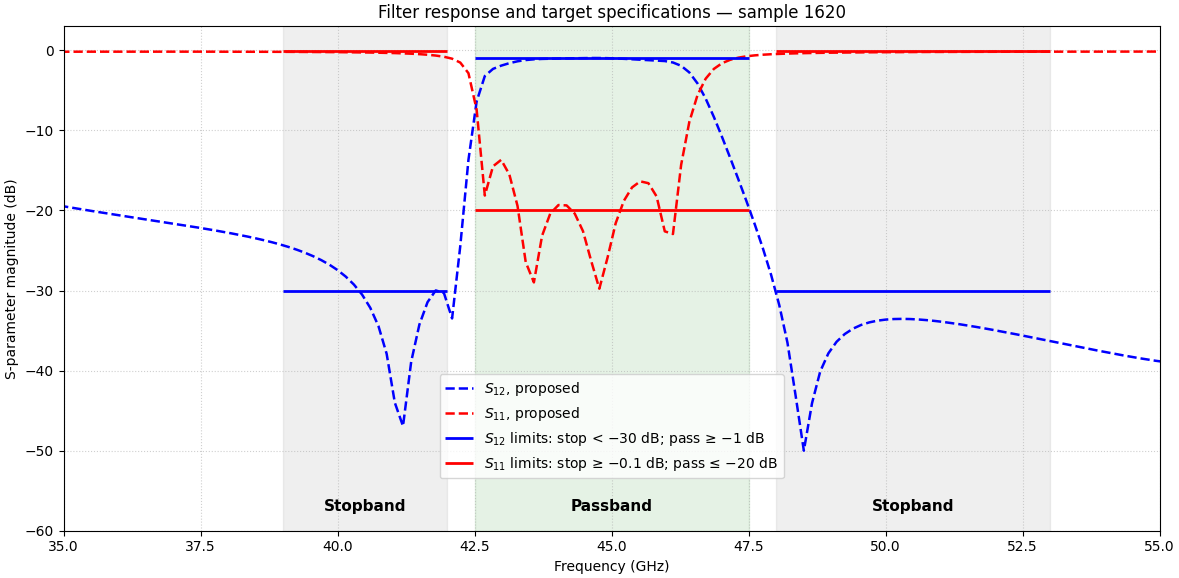}
\caption{Design from a specification after $1{,}620$ simulations: simulated $S_{21}$
(blue, dashed) and $S_{11}$ (red, dashed) of the best design, with the specification
limits (solid); passband shaded green, stopbands grey, transition bands excluded from the
loss.}
\label{fig:filter_spec}
\end{figure}

\paragraph{Design from a specification.} The surrogate predicts the real and imaginary
parts of $S_{11}$ and $S_{21}$ and is trained by the mean squared error between these
components and the CST responses. For design, let $T$ and $R$ denote the transmission
and reflection magnitudes in dB. The specification is written as signed violations,
positive when a limit is broken: $v=-1-T$ and $v=R+20$ at passband frequencies, and
$v=T+30$ and $v=-0.1-R$ at stopband frequencies. Each violation passes through a
softplus-smoothed leaky ReLU,
\begin{equation}
\rho(v)=\alpha v+\frac{1-\alpha}{\beta}\log\!\left(1+e^{\beta v}\right),
\label{eq:app_violation}
\end{equation}
which has slope $1$ for large violations and slope $\alpha$ for satisfied limits, so it
keeps rewarding margin, with $\alpha=0.01$ and $\beta=10\,\mathrm{dB}^{-1}$. The design
loss averages, with equal weights, the mean of $\rho$ over each of the four band-wise
terms (passband transmission and reflection, stopband transmission and reflection),
excluding the transition bands $42$--$42.5$ and $47.5$--$48$\,GHz. During the online
phase, each CST evaluation updates the surrogate, which is then frozen while the inverse
network is trained through this differentiable loss. Offline, the surrogate is retrained
on all accumulated samples, and the inverse network either with the achieved-response
objective~\eqref{eq:loss_d_offline} or with the same constraint loss for randomly
generated feasible specifications; the run in Figure~\ref{fig:filter_spec} uses the
achieved-response objective.

The specification is encoded as two target magnitude responses in dB, sampled on the
CST frequency grid and concatenated,
\begin{equation}
\mathbf{y}^{\star}=\left[T^{\star}(f_1),\ldots,T^{\star}(f_N),\,
R^{\star}(f_1),\ldots,R^{\star}(f_N)\right]\in\mathbb{R}^{2N},
\label{eq:app_spec_encoding}
\end{equation}
where $T^{\star}$ and $R^{\star}$ denote the target transmission and reflection
magnitudes. With frequency in GHz,
\begin{equation}
T^{\star}(f)=
\begin{cases}
-1, & 42.5\leq f\leq 47.5,\\
-30, & \text{otherwise},
\end{cases}
\qquad
R^{\star}(f)=
\begin{cases}
-20, & 42.5\leq f\leq 47.5,\\
-0.1, & \text{otherwise}.
\end{cases}
\label{eq:app_spec_targets}
\end{equation}
With $N=202$ frequency samples, the inverse network receives a $404$-dimensional input
and predicts normalised design parameters $\hat{\mathbf{x}}=D_{\theta}(\mathbf{y}^{\star})$.
The inequality directions are enforced by the specification loss: although the
transition-band entries of the input are populated, the loss evaluates only the passband
and the stopbands $f<42$ and $f>48$\,GHz.

After $1{,}620$ simulations, the best design (Figure~\ref{fig:filter_spec}) meets the
insertion-loss limit from about $43$ to $46.5$\,GHz, and the stopband limit over nearly
all of the upper stopband and the lower stopband above about $40.5$\,GHz. Below that its
rejection falls to about $20$\,dB at $35$\,GHz, and the passband edges and reflection
fall short in places.

\section{Offline objective for the inverse network}
\label{app:offline_target}

\begin{figure}[h]
\centering
\begin{subfigure}[b]{\textwidth}
  \includegraphics[width=\linewidth]{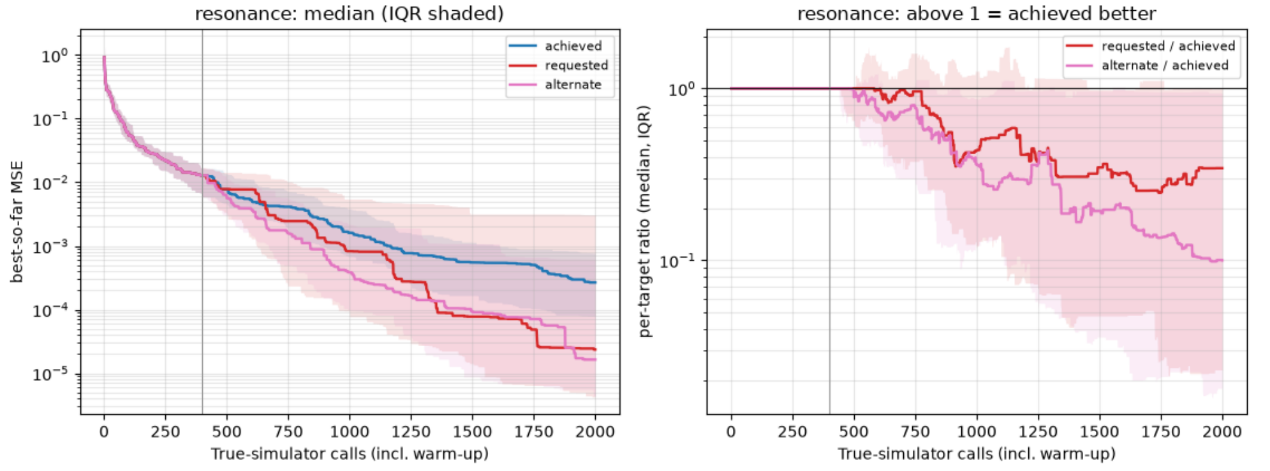}
  \caption{No exploration}
\end{subfigure}

\begin{subfigure}[b]{\textwidth}
  \includegraphics[width=\linewidth]{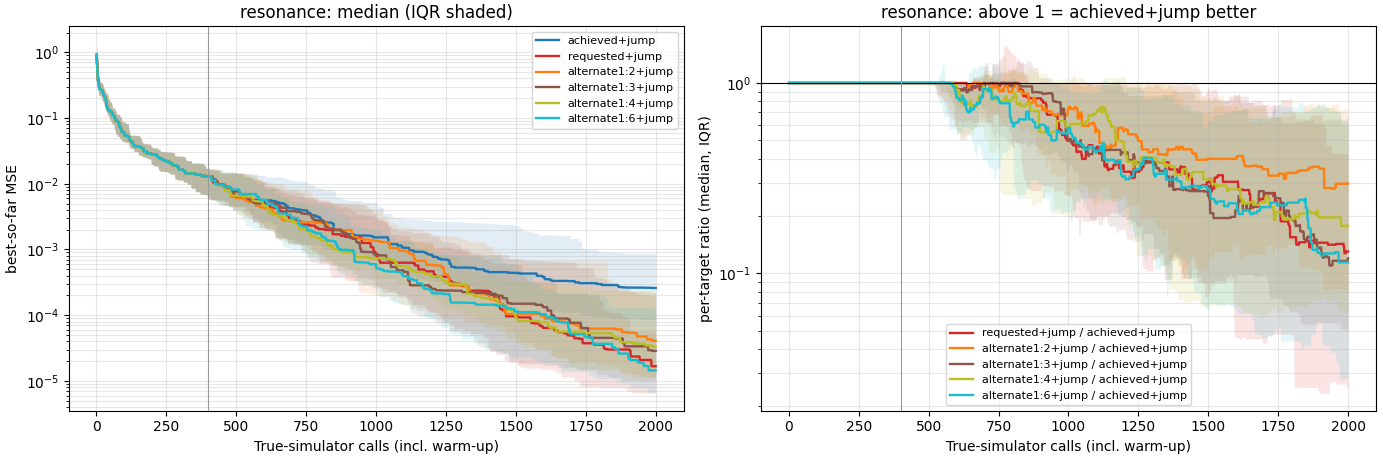}
  \caption{With exploration}
\end{subfigure}
\caption{Resonance: median best-so-far error (left) and per-target ratio to
achieved-response training (right; below $1$ = better), median and interquartile
range.}
\label{fig:app_target_res}
\end{figure}

\begin{figure}[h]
\centering
\begin{subfigure}[b]{\textwidth}
  \includegraphics[width=\linewidth]{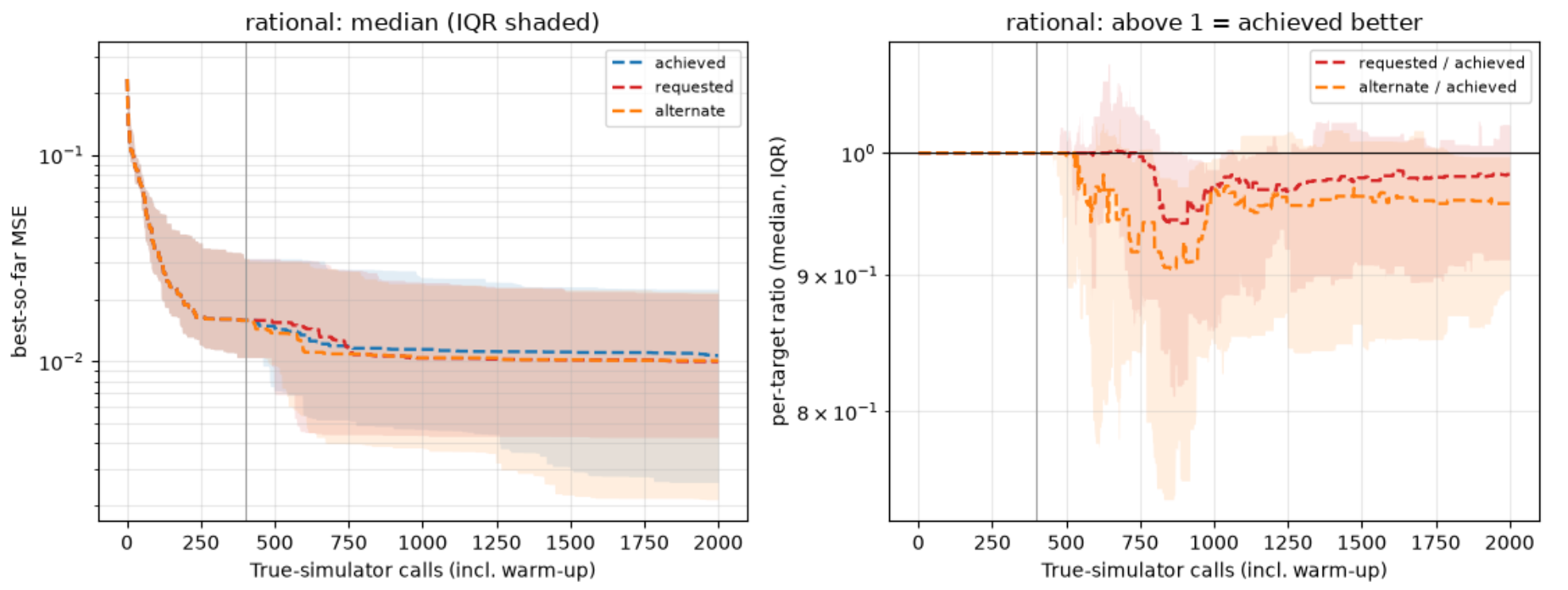}
  \caption{No exploration}
\end{subfigure}

\begin{subfigure}[b]{\textwidth}
  \includegraphics[width=\linewidth]{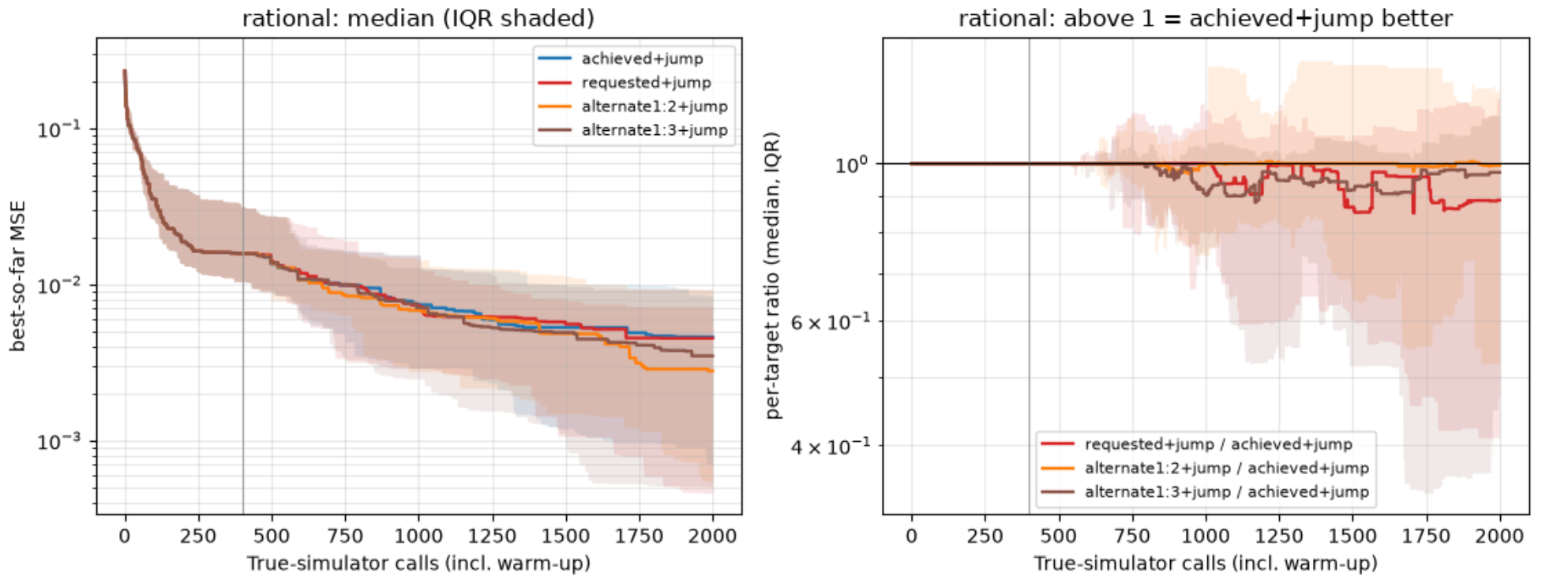}
  \caption{With exploration}
\end{subfigure}
\caption{Rational: as Figure~\ref{fig:app_target_res}.}
\label{fig:app_target_rat}
\end{figure}

The study of Section~\ref{sec:mechanism} uses $30$ targets per benchmark, a GA warm-up
of $400$ calls and $2{,}000$ calls in total. All variants of a benchmark share each
target's warm-up, pre-trained networks and random seeds, so they differ only in the
offline objective and, between settings, in exploration.
Figures~\ref{fig:app_target_res} and~\ref{fig:app_target_rat} show the median curves
and the per-target ratios to achieved-response training;
Table~\ref{tab:app_target_budgets} gives the paired comparisons at four budgets.
Comparisons at several budgets share targets and are correlated, so the final budget
is the primary test and the earlier ones describe the trajectory.

With exploration, variants could tie merely because they share the same exploratory
moves. On resonance this is not the case: most improvements come from ORO's own
proposals ($29$ to $36$ per target) rather than from exploratory calls ($14$ to $17$),
and counting only ORO's own proposals, whose outcome the offline objective controls,
gives the same conclusions: every variant with requested rounds beats achieved-response
training on $24$ to $26$ of $30$ targets, with per-target ratios of $0.12$ to $0.30$.

\begin{table}[h]
\centering
\caption{Offline objective for the inverse network: median final error after
$2{,}000$ calls ($30$ targets, GA warm-up of $400$), the per-target median ratio to
achieved-response training under the same exploration setting (below $1$ = better),
and the number of targets on which each variant beats achieved-response training.
Ratios $a{:}r$ alternate offline rounds between the achieved and requested targets. On
resonance, the no-exploration rows come from a separate run.}
\label{tab:offline_target}
\small
{
\begin{tabular}{llccc}
\toprule
 & Offline objective & Median & Ratio & Better than achieved \\
\midrule
\multirow{3}{*}{Resonance, no exploration}
 & achieved    & $2.67{\times}10^{-4}$ & -- & -- \\
 & requested   & $2.36{\times}10^{-5}$ & $0.34$ & $23/30$ ($p{=}5.2{\times}10^{-3}$) \\
 & $1{:}1$     & $1.64{\times}10^{-5}$ & $0.10$ & $23/30$ ($p{=}5.2{\times}10^{-3}$) \\
\midrule
\multirow{6}{*}{Resonance, exploration}
 & achieved    & $2.60{\times}10^{-4}$ & -- & -- \\
 & requested   & $1.67{\times}10^{-5}$ & $0.13$ & $27/30$ ($p{=}8.4{\times}10^{-6}$) \\
 & $1{:}2$     & $4.10{\times}10^{-5}$ & $0.31$ & $28/30$ ($p{=}8.7{\times}10^{-7}$) \\
 & $1{:}3$     & $2.87{\times}10^{-5}$ & $0.12$ & $26/30$ ($p{=}5.9{\times}10^{-5}$) \\
 & $1{:}4$     & $3.33{\times}10^{-5}$ & $0.18$ & $27/30$ ($p{=}8.4{\times}10^{-6}$) \\
 & $1{:}6$     & $1.45{\times}10^{-5}$ & $0.12$ & $26/30$ ($p{=}5.9{\times}10^{-5}$) \\
\midrule
\multirow{3}{*}{Rational, no exploration}
 & achieved    & $1.07{\times}10^{-2}$ & -- & -- \\
 & requested   & $9.97{\times}10^{-3}$ & $0.98$ & $19/30$ ($p{=}0.20$) \\
 & $1{:}1$     & $1.01{\times}10^{-2}$ & $0.96$ & $25/30$ ($p{=}3.2{\times}10^{-4}$) \\
\midrule
\multirow{4}{*}{Rational, exploration}
 & achieved    & $4.60{\times}10^{-3}$ & -- & -- \\
 & requested   & $4.52{\times}10^{-3}$ & $0.89$ & $16/29$ ($p{=}0.71$) \\
 & $1{:}2$     & $2.79{\times}10^{-3}$ & $0.99$ & $16/29$ ($p{=}0.71$) \\
 & $1{:}3$     & $3.48{\times}10^{-3}$ & $0.97$ & $19/30$ ($p{=}0.20$) \\
\bottomrule
\end{tabular}}
\end{table}

\begin{table}[h]
\centering
\caption{Targets on which each variant beats achieved-response training, and the
per-target median ratio to it, at four budgets (paired sign test, ties excluded and
listed). For $1{:}6$ the comparison is with $1{:}4$.}
\label{tab:app_target_budgets}
\scriptsize
{
\begin{tabular}{llcccc}
\toprule
 & Variant & $600$ calls & $1{,}000$ calls & $1{,}600$ calls & $2{,}000$ calls \\
\midrule
\multirow{2}{*}{Res., no expl.}
 & requested & $15/27$, 3 tied, $0.98$ & $21/30^{*}$, $0.44$ & $23/30^{**}$, $0.32$ & $23/30^{**}$, $0.34$ \\
 & $1{:}1$   & $20/29$, 1 tied, $0.76$ & $24/30^{**}$, $0.36$ & $22/30^{*}$, $0.20$ & $23/30^{**}$, $0.10$ \\
\midrule
\multirow{4}{*}{Res., expl.}
 & requested & $13/18$, 12 tied, $1.00$ & $21/29^{*}$, 1 tied, $0.50$ & $24/30^{**}$, $0.32$ & $27/30^{***}$, $0.13$ \\
 & $1{:}2$   & $17/19^{***}$, 11 tied, $0.91$ & $20/28^{*}$, 2 tied, $0.73$ & $26/30^{***}$, $0.40$ & $28/30^{***}$, $0.31$ \\
 & $1{:}3$   & $17/20^{**}$, 10 tied, $0.94$ & $22/28^{**}$, 2 tied, $0.51$ & $26/30^{***}$, $0.22$ & $26/30^{***}$, $0.12$ \\
 & $1{:}6$ vs.\ $1{:}4$ & $13/23$, 7 tied, $1.00$ & $13/29$, 1 tied, $1.02$ & $13/29$, 1 tied, $1.01$ & $17/29$, 1 tied, $0.85$ \\
\midrule
\multirow{2}{*}{Rat., no expl.}
 & requested & $14/25$, 5 tied, $1.00$ & $24/29^{***}$, 1 tied, $0.97$ & $20/30$, $0.98$ & $19/30$, $0.98$ \\
 & $1{:}1$   & $17/27$, 3 tied, $0.97$ & $24/30^{**}$, $0.97$ & $21/30^{*}$, $0.96$ & $25/30^{***}$, $0.96$ \\
\midrule
\multirow{3}{*}{Rat., expl.}
 & requested & $5/15$, 15 tied, $1.00$ & $13/26$, 4 tied, $1.00$ & $17/28$, 2 tied, $0.97$ & $16/29$, 1 tied, $0.89$ \\
 & $1{:}2$   & $7/14$, 16 tied, $1.00$ & $15/27$, 3 tied, $1.00$ & $14/28$, 2 tied, $1.00$ & $16/29$, 1 tied, $0.99$ \\
 & $1{:}3$   & $10/17$, 13 tied, $1.00$ & $18/29$, 1 tied, $0.91$ & $22/30^{*}$, $0.91$ & $19/30$, $0.97$ \\
\bottomrule
\end{tabular}}

\smallskip
{\scriptsize $^{*}p<0.05$, $^{**}p<0.01$, $^{***}p<0.001$ (two-sided sign test).}
\end{table}

\section{Gradient alignment}
\label{app:gradients}

At regular probes during refinement we compute, at the current proposal, the gradient
of the target error $\ell(S_\phi(x),y^{\star})$ through the surrogate and the true
gradient of $\ell(F(x),y^{\star})$, obtained by automatic differentiation of the
synthetic simulator. True gradients are used only for this measurement, never for
optimization. We record their cosine in box-normalised coordinates, before and after
the step's online correction; whether a small step along the negative surrogate
gradient reduces the true error (a descent direction); the ratio of gradient norms;
and the cosine between the resulting updates of the inverse network's weights. Probes
cover $10$ targets with $160$ probes each per variant and benchmark. ``No offline''
removes the offline phase and keeps everything else. These runs use the
achieved-response offline objective~\eqref{eq:loss_d_offline} and no exploration.

\begin{table}[h]
\centering
\caption{Surrogate versus true gradients: medians over all probes. ``Before'' and
``after'' refer to the online correction of the surrogate in the same step.}
\label{tab:app_gradients}
\small
{
\begin{tabular}{llcccccc}
\toprule
 & & \multicolumn{2}{c}{Cosine} & \multicolumn{2}{c}{Descent direction} & Norm ratio & Weight-update \\
\cmidrule(lr){3-4}\cmidrule(lr){5-6}
Benchmark & Variant & before & after & before & after & $|g_S|/|g_F|$ & cosine \\
\midrule
\multirow{2}{*}{Resonance} & ORO        & $0.73$ & $0.77$ & $90\%$ & $92\%$ & $0.61$ & $0.79$ \\
                           & No offline & $-0.02$ & $-0.02$ & $47\%$ & $47\%$ & $0.38$ & $0.15$ \\
\midrule
\multirow{2}{*}{Rational}  & ORO        & $0.38$ & $0.43$ & $77\%$ & $79\%$ & $0.36$ & $0.44$ \\
                           & No offline & $-0.02$ & $-0.02$ & $46\%$ & $47\%$ & $0.50$ & $0.01$ \\
\bottomrule
\end{tabular}}
\end{table}

\begin{figure}[h]
\centering
\includegraphics[width=0.85\linewidth]{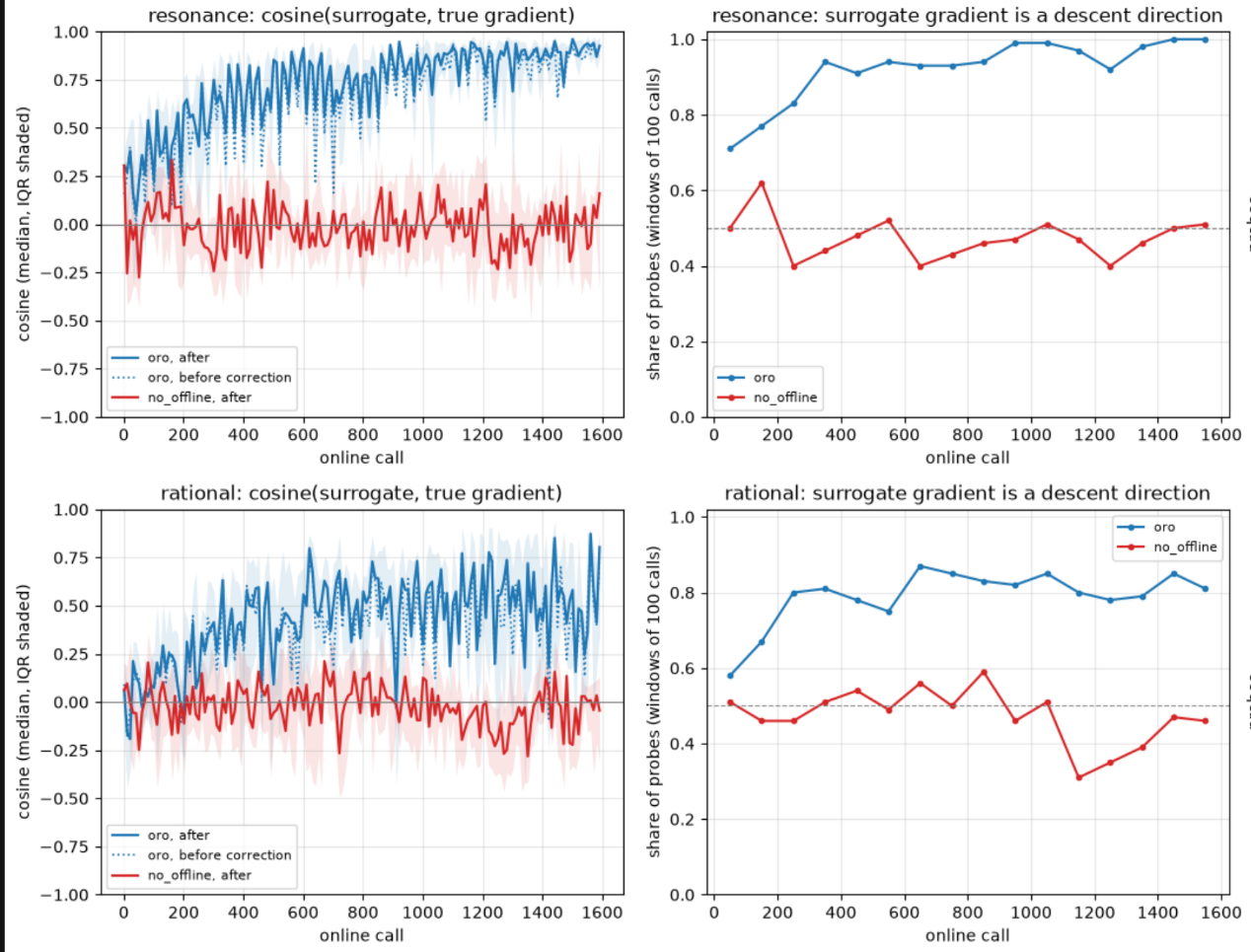}
\caption{Cosine between surrogate and true gradients (left; median and
interquartile range) and the share of probes where the surrogate gradient is a descent
direction for the true error (right; windows of $100$ calls), with and without the
offline phase.}
\label{fig:app_gradients}
\end{figure}

\section{Inverse network versus direct design optimization}
\label{app:direct_design}

\begin{figure}[h]
\centering
\begin{subfigure}[b]{\textwidth}
  \includegraphics[width=\linewidth]{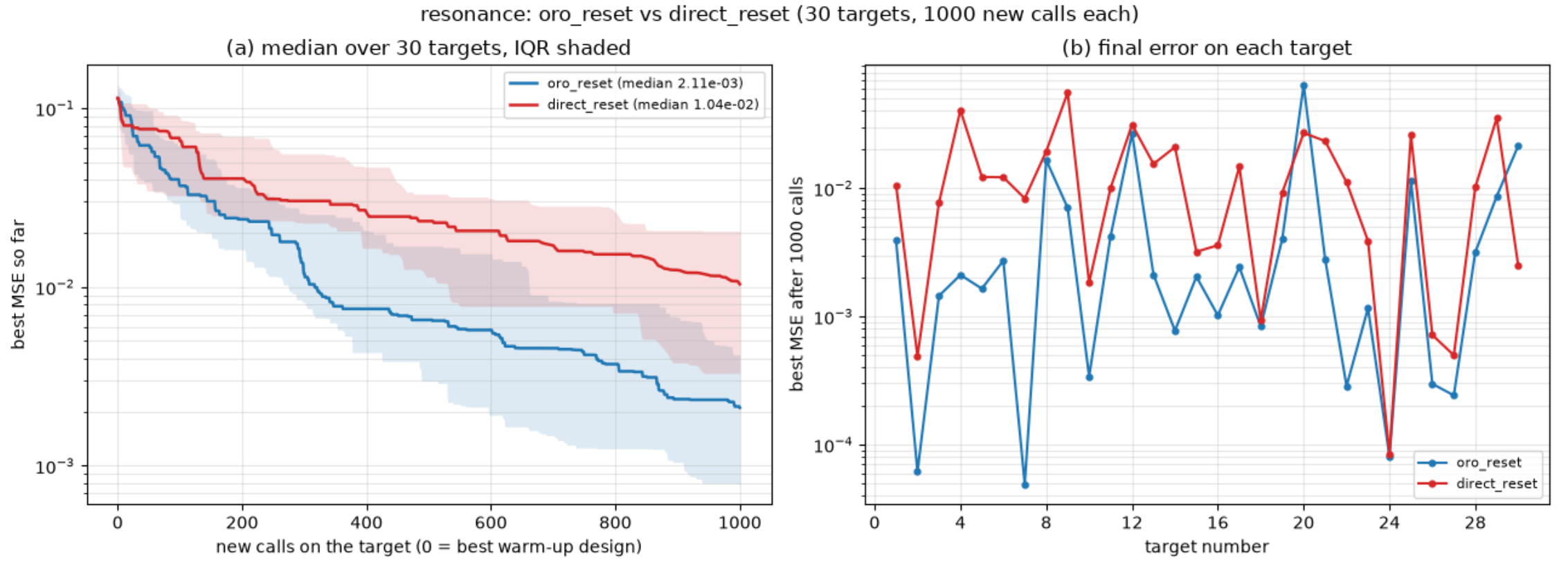}
  \caption{Resonance, $1{,}000$ new calls per target}
\end{subfigure}

\begin{subfigure}[b]{\textwidth}
  \includegraphics[width=\linewidth]{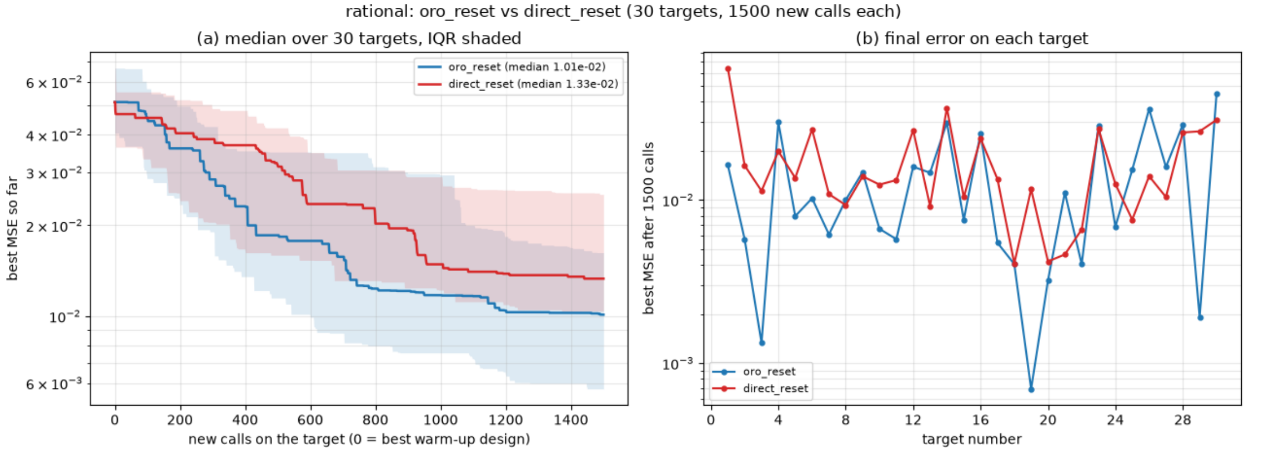}
  \caption{Rational, $1{,}500$ new calls per target}
\end{subfigure}
\caption{ORO versus direct design optimization: median best-so-far error with
interquartile range (left) and final error on each target (right).}
\label{fig:app_direct_design}
\end{figure}

Direct design optimization keeps ORO's surrogate, online correction, offline
retraining, learning rates and schedule, but replaces the inverse network by a single
design vector updated by gradient steps through the corrected surrogate. Both start
each target from the same Latin-hypercube warm-up of $400$ designs and pre-trained
surrogate. Table~\ref{tab:app_direct_design} gives the paired comparison and
Figure~\ref{fig:app_direct_design} the curves.
Both use the achieved-response offline objective~\eqref{eq:loss_d_offline} and no
exploration.

\begin{table}[h]
\centering
\caption{ORO versus direct design optimization through the same corrected
surrogate: median best-so-far error after a given number of new calls, and the number
of targets on which ORO is better (paired sign test, ties excluded).}
\label{tab:app_direct_design}
\small
{
\begin{tabular}{llccc}
\toprule
Benchmark & Calls & ORO & Direct design & ORO better \\
\midrule
\multirow{4}{*}{Resonance}
 & $50$      & $6.21{\times}10^{-2}$ & $7.65{\times}10^{-2}$ & $16/30$ ($p{=}0.86$) \\
 & $100$     & $3.69{\times}10^{-2}$ & $6.59{\times}10^{-2}$ & $20/30$ ($p{=}0.099$) \\
 & $300$     & $1.20{\times}10^{-2}$ & $3.03{\times}10^{-2}$ & $24/30$ ($p{=}1.4{\times}10^{-3}$) \\
 & $1{,}000$ & $\mathbf{2.11{\times}10^{-3}}$ & $1.04{\times}10^{-2}$ & $28/30$ ($p{=}8.7{\times}10^{-7}$) \\
\midrule
\multirow{4}{*}{Rational}
 & $50$      & $5.12{\times}10^{-2}$ & $4.68{\times}10^{-2}$ & $5/21$, 9 tied ($p{=}0.027$) \\
 & $100$     & $4.54{\times}10^{-2}$ & $4.55{\times}10^{-2}$ & $8/25$, 5 tied ($p{=}0.11$) \\
 & $300$     & $2.84{\times}10^{-2}$ & $3.86{\times}10^{-2}$ & $17/29$, 1 tied ($p{=}0.46$) \\
 & $1{,}500$ & $\mathbf{1.01{\times}10^{-2}}$ & $1.33{\times}10^{-2}$ & $18/30$ ($p{=}0.36$) \\
\bottomrule
\end{tabular}}
\end{table}

\section{Sequential design}
\label{app:sequential}

Thirty targets are designed in sequence with $1{,}000$ new calls each. ORO's
Latin-hypercube warm-up of $400$ calls is paid once, so over the sequence ORO spends
$30{,}400$ calls against $30{,}000$ for each baseline. Between targets both networks are
re-initialised and pre-trained on the whole history for $3{,}000$ epochs; within a target,
the offline phase uses the $N_{\text{sel}}$ designs with the lowest error for that target
($800$ on rational, $2{,}500$ on resonance and Rosenbrock), the inverse network trains on the requested
target offline, and every second call is exploratory, so $500$ calls per target explore,
of which $12.1$ on resonance, $33.8$ on rational and $0.6$ on Rosenbrock improved on
every earlier design. TuRBO's GP fits failed in $6$ of $3{,}565$ fits on resonance and
none of $3{,}551$ on rational or of $3{,}556$ on Rosenbrock. Table~\ref{tab:sequential} gives the final
errors and Table~\ref{tab:app_seq_trend} the trend over the sequence; the median curves
are in Figure~\ref{fig:sequential}.

\begin{table}[h]
\centering
\caption{Sequential design of $30$ targets with $1{,}000$ new calls each: median and
mean final error, and the number of targets on which ORO is better (paired two-sided
sign test). ORO pays for one warm-up of $400$ calls in addition; every baseline starts
each target from scratch.}
\label{tab:sequential}
\footnotesize
{
\begin{tabular}{lccc}
\toprule
Method & Median & Mean & ORO better \\
\midrule
\multicolumn{4}{l}{\textit{Resonance}} \\
ORO          & $\mathbf{2.44{\times}10^{-5}}$ & $\mathbf{9.70{\times}10^{-4}}$ & -- \\
TuRBO        & $4.98{\times}10^{-4}$ & $2.81{\times}10^{-3}$ & $28/30$ ($p{=}8.7{\times}10^{-7}$) \\
CMA-ES       & $2.42{\times}10^{-2}$ & $2.48{\times}10^{-2}$ & $30/30$ ($p{=}1.9{\times}10^{-9}$) \\
GA           & $3.44{\times}10^{-3}$ & $5.63{\times}10^{-3}$ & $29/30$ ($p{=}5.8{\times}10^{-8}$) \\
BO           & $5.59{\times}10^{-2}$ & $5.66{\times}10^{-2}$ & $30/30$ ($p{=}1.9{\times}10^{-9}$) \\
AutoTandemML & $1.01{\times}10^{-1}$ & $9.93{\times}10^{-2}$ & $30/30$ ($p{=}1.9{\times}10^{-9}$) \\
\midrule
\multicolumn{4}{l}{\textit{Rosenbrock}} \\
ORO          & $\mathbf{1.17{\times}10^{-6}}$ & $\mathbf{2.74{\times}10^{-6}}$ & -- \\
TuRBO        & $3.28{\times}10^{-4}$ & $4.39{\times}10^{-4}$ & $30/30$ ($p{=}1.9{\times}10^{-9}$) \\
CMA-ES       & $2.40{\times}10^{-3}$ & $2.84{\times}10^{-3}$ & $30/30$ ($p{=}1.9{\times}10^{-9}$) \\
GA           & $7.60{\times}10^{-3}$ & $8.92{\times}10^{-3}$ & $30/30$ ($p{=}1.9{\times}10^{-9}$) \\
BO           & $1.43{\times}10^{-2}$ & $1.72{\times}10^{-2}$ & $30/30$ ($p{=}1.9{\times}10^{-9}$) \\
AutoTandemML & $3.28{\times}10^{-2}$ & $3.64{\times}10^{-2}$ & $30/30$ ($p{=}1.9{\times}10^{-9}$) \\
\midrule
\multicolumn{4}{l}{\textit{Rational}} \\
ORO          & $\mathbf{3.17{\times}10^{-3}}$ & $\mathbf{5.04{\times}10^{-3}}$ & -- \\
TuRBO        & $4.90{\times}10^{-3}$ & $5.81{\times}10^{-3}$ & $18/30$ ($p{=}0.36$) \\
CMA-ES       & $2.55{\times}10^{-2}$ & $2.51{\times}10^{-2}$ & $30/30$ ($p{=}1.9{\times}10^{-9}$) \\
GA           & $1.06{\times}10^{-2}$ & $1.63{\times}10^{-2}$ & $26/30$ ($p{=}5.9{\times}10^{-5}$) \\
BO           & $2.09{\times}10^{-2}$ & $2.29{\times}10^{-2}$ & $30/30$ ($p{=}1.9{\times}10^{-9}$) \\
AutoTandemML & $4.89{\times}10^{-2}$ & $4.85{\times}10^{-2}$ & $30/30$ ($p{=}1.9{\times}10^{-9}$) \\
\bottomrule
\end{tabular}}
\end{table}

\begin{table}[h]
\centering
\caption{Trend over the sequence: Spearman correlation between target index and
ORO's best error after a given number of new calls ($p$ in parentheses). Negative values
mean later targets reach lower error.}
\label{tab:app_seq_trend}
\footnotesize
{
\resizebox{\linewidth}{!}{%
\begin{tabular}{lccccc}
\toprule
Benchmark & $0$ calls & $50$ & $100$ & $300$ & $1{,}000$ \\
\midrule
Resonance & $-0.46$ ($0.011$) & $-0.10$ ($0.61$) & $-0.11$ ($0.56$) & $-0.54$ ($0.0019$) & $-0.17$ ($0.38$) \\
Rosenbrock & $-0.51$ ($0.0042$) & $-0.83$ ($1.2{\times}10^{-8}$) & $-0.61$ ($3.7{\times}10^{-4}$) & $-0.80$ ($1.1{\times}10^{-7}$) & $-0.87$ ($2.6{\times}10^{-10}$) \\
Rational  & $-0.31$ ($0.097$) & $-0.31$ ($0.10$) & $-0.30$ ($0.11$) & $-0.26$ ($0.17$) & $-0.19$ ($0.31$) \\
\bottomrule
\end{tabular}}}
\end{table}

\section{Wall-clock time}
\label{app:wallclock}

Table~\ref{tab:app_wallclock} gives the mean wall-clock time per target in the
budget-matched comparison, on one GPU. ORO's pre-training on the warm-up set is listed
separately, and its warm-up is GA's first $400$ calls;
AutoTandemML's one-off dataset collection and training are divided by the number of
targets. The simulators here are cheap, so these times measure each method's own
overhead. With a full-wave solver they are small by comparison: at about $5$ minutes per
CST simulation of our filter, the $2{,}800$ calls of one target take $840{,}000$\,s ($233$ hours), over
$1{,}900$ times ORO's largest overhead in Table~\ref{tab:app_wallclock}.

\begin{table}[h]
\centering
\caption{Wall-clock seconds per target at $2{,}800$ calls. For ORO, the breakdown
is online refinement (of which simulator calls), offline surrogate training and offline
inverse-network training. Pre-training was not timed: its row is estimated from the
measured cost per update of the offline phases, as pre-training ($2{,}500$ epochs over
the $400$ warm-up designs) performs $10{,}000$ updates per network, $6.4\%$ of the
$156{,}090$ of a run's offline phases.}
\label{tab:app_wallclock}
\small
{
\begin{tabular}{lccc}
\toprule
Method & Resonance & Rosenbrock & Rational \\
\midrule
GA            & $0.3$ & $0.3$ & $0.3$ \\
CMA-ES        & $0.2$ & $0.2$ & $0.3$ \\
AutoTandemML  & $15.3$ & $16.2$ & $15.5$ \\
BO            & $111.4$ & $111.1$ & $400.7$ \\
TuRBO         & $113.2$ & $97.9$ & $343.0$ \\
ORO           & $336.1$ & $389.3$ & $425.4$ \\
\quad online (simulator)      & $7.7$ ($1.0$) & $8.2$ ($0.9$) & $9.6$ ($1.7$) \\
\quad offline, surrogate      & $127.6$ & $149.2$ & $162.6$ \\
\quad offline, inverse network & $200.8$ & $232.0$ & $253.2$ \\
ORO pre-training (once per target), estimated & ${\approx}21$ & ${\approx}24$ & ${\approx}27$ \\
\bottomrule
\end{tabular}}
\end{table}

\section{Example targets}
\label{app:examples}

\begin{figure}[h]
\centering
\begin{subfigure}[b]{0.49\textwidth}\includegraphics[width=\linewidth]{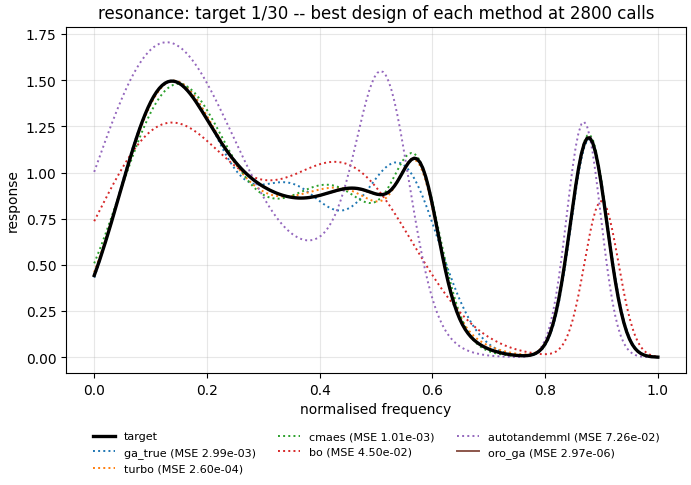}\end{subfigure}
\hfill
\begin{subfigure}[b]{0.49\textwidth}\includegraphics[width=\linewidth]{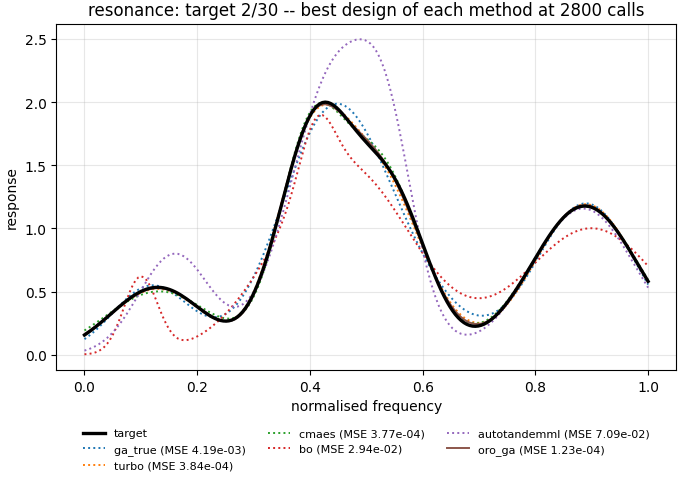}\end{subfigure}

\begin{subfigure}[b]{0.49\textwidth}\includegraphics[width=\linewidth]{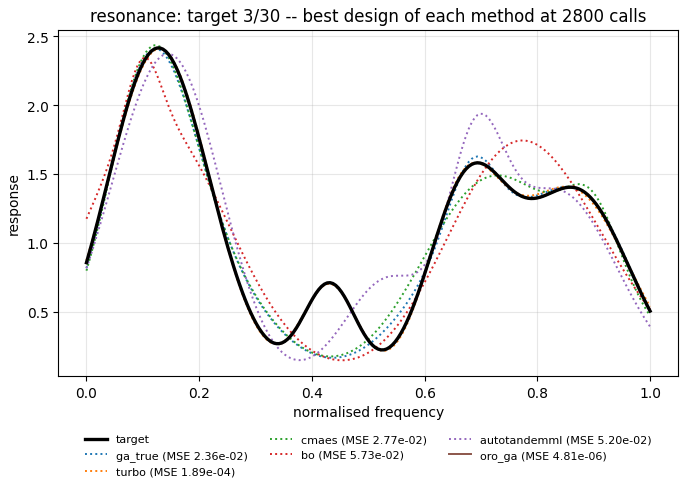}\end{subfigure}
\hfill
\begin{subfigure}[b]{0.49\textwidth}\includegraphics[width=\linewidth]{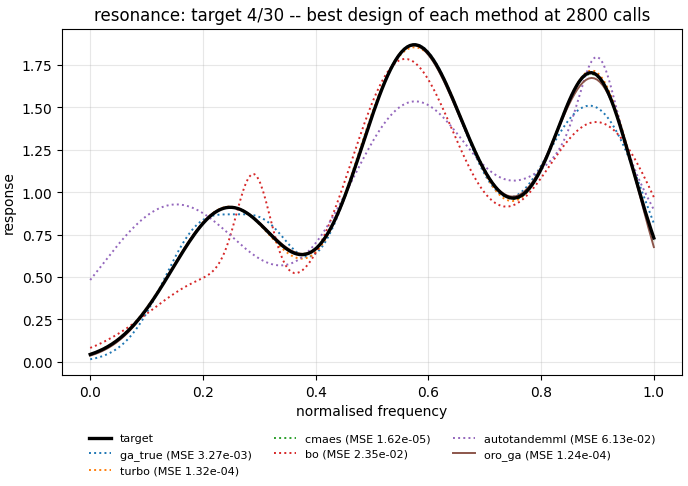}\end{subfigure}
\caption{Resonance: target response (black) and the best design of each method
at $2{,}800$ calls; errors in the legends.}
\label{fig:app_examples_res}
\end{figure}

\begin{figure}[h]
\centering
\begin{subfigure}[b]{0.49\textwidth}\includegraphics[width=\linewidth]{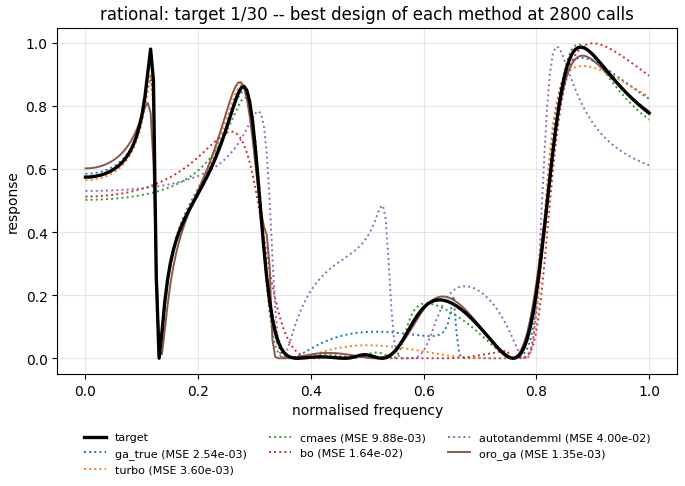}\end{subfigure}
\hfill
\begin{subfigure}[b]{0.49\textwidth}\includegraphics[width=\linewidth]{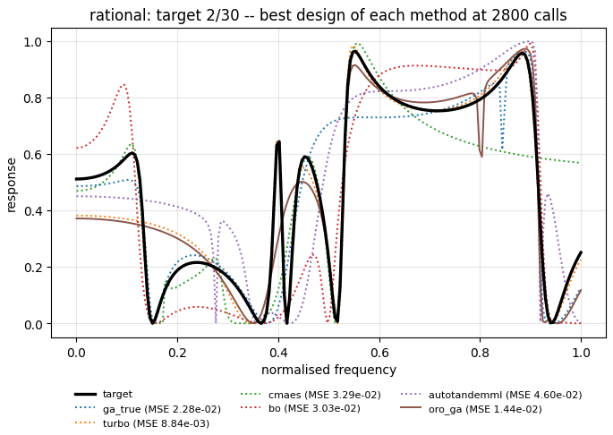}\end{subfigure}

\begin{subfigure}[b]{0.49\textwidth}\includegraphics[width=\linewidth]{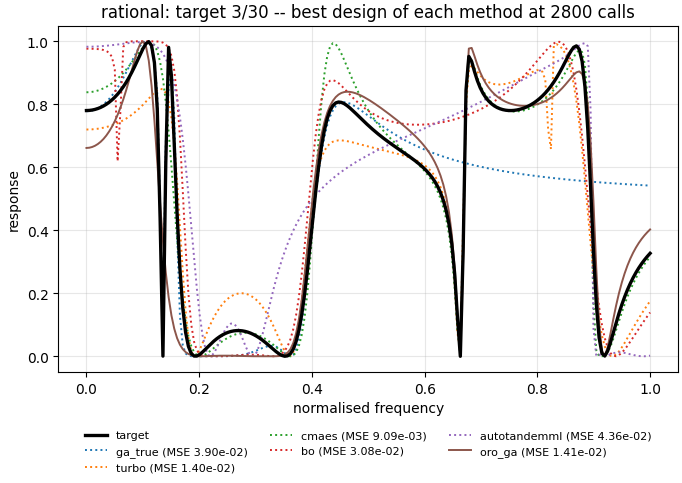}\end{subfigure}
\hfill
\begin{subfigure}[b]{0.49\textwidth}\includegraphics[width=\linewidth]{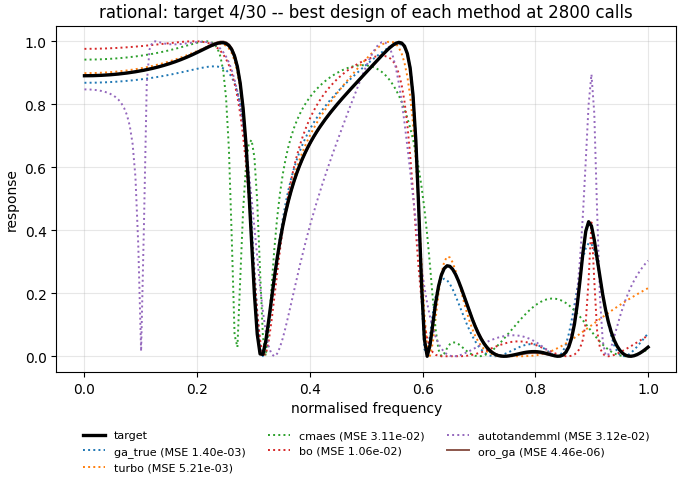}\end{subfigure}
\caption{Rational: as Figure~\ref{fig:app_examples_res}.}
\label{fig:app_examples_rat}
\end{figure}

Figures~\ref{fig:app_examples_res} and~\ref{fig:app_examples_rat} show the first four
targets of each benchmark with the best design found by each method at $2{,}800$
calls.

\end{document}